\documentclass[]{opendatalab}
\usepackage{xcolor}
\usepackage{longtable}
\usepackage{tabularx}
\usepackage{listings}
\usepackage{tabularx}
\usepackage{amsmath}
\usepackage{xltabular}
\usepackage{booktabs}
\usepackage{ragged2e}
\usepackage{array}
\usepackage{booktabs}
\usepackage{inconsolata}
\usepackage[numbers]{natbib}
\usepackage[textsize=tiny]{todonotes}
\usepackage{setspace}     
\usepackage[export]{adjustbox}
\usepackage{threeparttable}
\usepackage{multirow}
\usepackage{enumitem}
\usepackage{newfloat}
\usepackage{listings}
\usepackage{colortbl}
\usepackage{amsfonts}
\usepackage{amssymb}
\usepackage{pifont}
\usepackage{float}
\usepackage{bm}
\usepackage{pifont}
\usepackage{fbox}
\usepackage{tabulary}
\usepackage{makecell}
\usepackage{bbm}
\usepackage{mdframed}
\usepackage{algorithm}
\usepackage{algorithmic}
\usepackage{caption,subcaption}
\usepackage{adjustbox}
\usepackage{array}
\usepackage{booktabs}
\usepackage{tcolorbox}
\usepackage{graphicx}
\usepackage{wrapfig}

\tcbuselibrary{theorems}
\definecolor{lightgray}{gray}{.9}
\definecolor{deepgray}{gray}{.8}
\tcbset{highlight math/.append style={left=0mm,right=0mm,top=0mm,bottom=0mm, colframe=white}}

\newcolumntype{I}{!{\vrule width 1pt}}
\makeatletter
\newcommand{\thickhline}{%
    \noalign {\ifnum 0=`}\fi \hrule height 1pt
    \futurelet \reserved@a \@xhline
}

\newcommand{\arr}{\ \textcolor{gray}{$\to$}\ \allowbreak}
\newcommand{\slb}{/\allowbreak}

\crefname{proposition}{Prop.}{Props.}
\crefname{section}{Sec.}{Secs.}
\crefname{table}{Tab.}{Tabs.}

\usepackage{xspace}
\makeatletter
\DeclareRobustCommand\onedot{\futurelet\@let@token\@onedot}
\def\@onedot{\ifx\@let@token.\else.\null\fi\xspace}

\usepackage[dvipsnames]{xcolor}
\usepackage{colortbl}
\definecolor{ada_blue}{rgb}{0,205,205}
\definecolor{glt_red}{rgb}{109,205,255}
\definecolor{MorandiBlue}{RGB}{118,134,146}
\definecolor{demphcolor}{RGB}{144,144,144}
\definecolor{mygray}{gray}{0.4}
\definecolor{autopurple}{HTML}{7030A0}
\definecolor{dyna_yellow}{HTML}{BF9000}
\definecolor{adaptive_blue}{HTML}{0070C0}
\definecolor{darkgrey}{RGB}{120,120,120}
\definecolor{mygrey}{RGB}{200,200,200}
\definecolor{myblue}{HTML}{00CDCD}
\definecolor{champagne}{rgb}{0.97, 0.91, 0.81}
\definecolor{darksalmon}{rgb}{0.91, 0.59, 0.48}
\definecolor{emerald}{rgb}{0.31, 0.78, 0.47}
\definecolor{green(pigment)}{rgb}{0.0, 0.65, 0.31}
\definecolor{amaranth}{rgb}{0.9, 0.17, 0.31}
\definecolor{iris}{rgb}{0.35, 0.31, 0.81}
\definecolor{uu}{rgb}{0.95, 0.51, 0.51}
\definecolor{spirodiscoball}{rgb}{0.06, 0.75, 0.99}
\definecolor{cadetblue}{RGB}{95,158,160} 
\definecolor{keywordcolor}{RGB}{178,34,34} 

\definecolor{customgreen}{HTML}{667b5b}

\definecolor{customblue}{HTML}{bcccea}

\usepackage{pifont}
\usepackage{bbding}

\definecolor{codegreen}{rgb}{0,0.6,0}
\definecolor{codegray}{rgb}{0.5,0.5,0.5}
\definecolor{codepurple}{rgb}{0.58,0,0.82}
\definecolor{backcolour}{rgb}{0.95,0.95,0.92}
\definecolor{promptcolor}{HTML}{D1D0F2}
\definecolor{promptcolorheader}{HTML}{bdbcec}
\newcommand{\promptbox}[2]{
\begin{tcolorbox}[
top=0.3em,bottom=0.3em,left=0.5em,right=0.5em,
toptitle=0.3em,bottomtitle=0.2em,boxsep=0pt,
colframe=promptcolorheader,colback=promptcolor!50,boxrule=0.5pt,
]
\footnotesize

\end{tcolorbox}
}
\lstdefinestyle{mystyle}{
    backgroundcolor=\color{backcolour},   
    commentstyle=\color{codegreen},
    keywordstyle=\color{magenta},
    numberstyle=\tiny\color{codegray},
    stringstyle=\color{codepurple},
    basicstyle=\ttfamily\footnotesize,
    breakatwhitespace=false,         
    breaklines=true,                 
    captionpos=b,                    
    keepspaces=true,                 
    numbers=left,                    
    numbersep=5pt,                  
    showspaces=false,                
    showstringspaces=false,
    showtabs=false,                  
    tabsize=2
}

\title{Tracing the Roots: A Multi-Agent Framework for Uncovering Data Lineage in Post-Training LLMs}

\author[1,2]{Yu Li}
\author[1]{Xiaoran Shang}
\author[1]{Qizhi Pei}
\author[1]{Yun Zhu}
\author[1,3]{Xin Gao}
\author[1,3]{Honglin Lin}
\author[1,3]{Zhanping Zhong}
\author[1]{Zhuoshi Pan}
\author[1]{Zheng Liu}
\author[1]{Xiaoyang Wang}
\author[1]{Conghui He}
\author[1]{Dahua Lin}
\author[2]{Feng Zhao}
\author[1]{Lijun Wu}

\affiliation[1]{Shanghai Artificial Intelligence Laboratory, OpenDataLab}
\affiliation[2]{University of Science and Technology of China}
\affiliation[3]{Shanghai Jiao Tong University}

\abstract{
Post-training data plays a pivotal role in shaping the capabilities of Large Language Models (LLMs), yet datasets are often treated as isolated artifacts, overlooking the systemic connections that underlie their evolution. To disentangle these complex relationships, we introduce the concept of \textbf{data lineage} to the LLM ecosystem and propose an automated multi-agent framework to reconstruct the evolutionary graph of dataset development. Through large-scale lineage analysis, we characterize domain-specific structural patterns, such as vertical refinement in math-oriented datasets and horizontal aggregation in general-domain corpora. Moreover, we uncover pervasive systemic issues, including \textit{structural redundancy} induced by implicit dataset intersections and the \textit{propagation of benchmark contamination} along lineage paths. To demonstrate the practical value of lineage analysis for data construction, we leverage the reconstructed lineage graph to create a \textit{lineage-aware diversity-oriented dataset}. By anchoring instruction sampling at upstream root sources, this approach mitigates downstream homogenization and hidden redundancy, yielding a more diverse post-training corpus. We further highlight lineage-centric analysis as an efficient and robust topological alternative to sample-level dataset comparison for large-scale data ecosystems. By grounding data construction in explicit lineage structures, our work advances post-training data curation toward a more systematic and controllable paradigm.
}

\date{\today}
\correspondence{Lijun Wu, \email{wulijun@pjlab.org.cn}}
\metadata[Homepage]{\url{https://arena.opendatalab.org.cn/data-lineage/website/index.html}}
\metadata[Code]{\url{https://github.com/Leey21/data-lineage}}

\begin{document}

\vspace{-0.3cm}
\begin{figure}[th]
    \hspace{5cm}
    \includegraphics[width=0.3\linewidth]{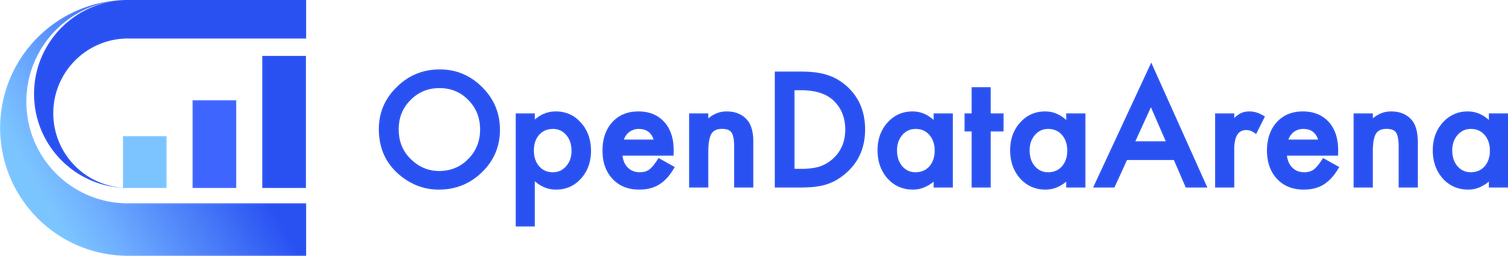}
    \label{fig:logo}
\end{figure}
\vspace{-0.2cm}

\maketitle

\section{Introduction}

High-quality post-training data is the primary engine driving LLM capabilities~\cite{zhou2023lima,cai2025opendataarenafairopenarena}, yet the community lacks systematic mechanisms to track its provenance. While recent efforts have extensively traced the evolution of model architectures~\cite{sajjadi2025survey,zhao2025surveylargelanguagemodels}, datasets are still predominantly treated as isolated artifacts, obscuring their true developmental context. In practice, modern post-training corpora are rarely constructed from scratch; instead, they emerge through recursive derivation processes that repurpose existing resources via semantic evolution~\cite{xu2024wizardlm}, knowledge distillation~\cite{mitra2024orcamath}, and structured fusion~\cite{pei2025mathfusion}. As a result, post-training datasets collectively form a dense and interdependent evolutionary network whose connections are largely undocumented.

\begin{figure}[t]
\centering
\includegraphics[width=0.86\linewidth, trim=3.5cm 0.7cm 4cm 1cm, clip]{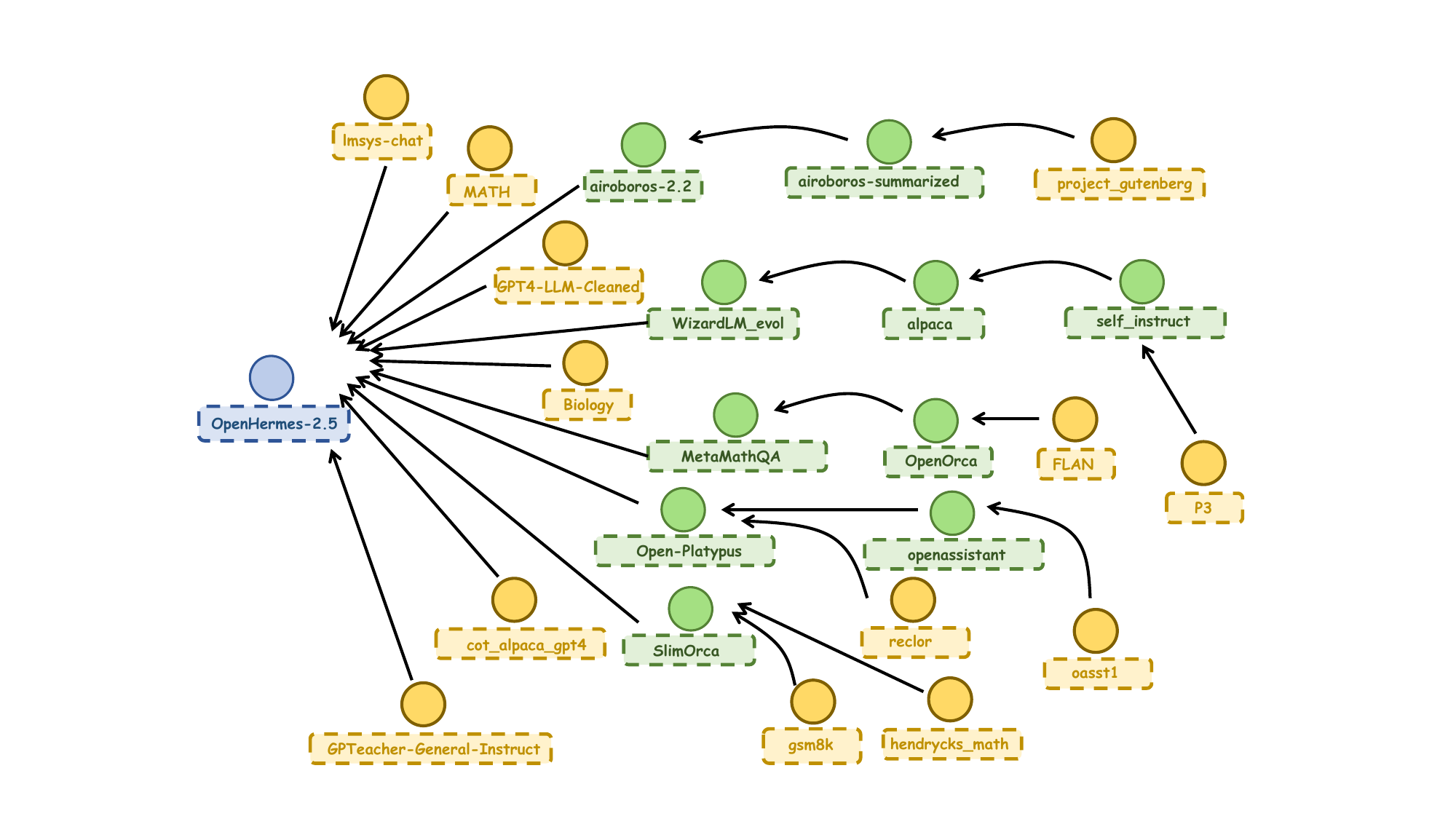}
\caption{Lineage graph construction with depth 3, using the OpenHermes-2.5 dataset~\cite{OpenHermes2.5} as an example. Yellow and green nodes denote leaf and internal nodes, respectively.}
\label{fig:case1}
\vspace{-5mm}
\end{figure}

This lack of lineage transparency raises two risks.
First, \textit{structural redundancy} emerges when datasets implicitly inherit overlapping sources, causing downstream corpora to converge semantically despite apparent scale growth~\cite{zhou2025survey}. Such hidden intersections erode effective diversity and weaken the marginal value of additional data.
Second, the \textit{propagation of benchmark contamination} becomes unavoidable when test samples embedded in upstream datasets are unknowingly inherited by downstream derivatives, introducing latent leakage that undermines the credibility of future evaluations~\cite{sainz2023nlp}. Without explicit lineage awareness, these risks remain difficult to detect or mitigate at scale.

To address this challenge, we introduce the concept of \textbf{data lineage} and propose a multi-agent collaborative framework that autonomously mines unstructured documentation and performs self-verified provenance tracing. Starting from 83 high-impact seed datasets spanning four major domains, we construct an evolutionary graph comprising 430 unique nodes connected by 971 inheritance edges (a real visualization example is shown in Figure~\ref{fig:case1}).
We analyze this ecosystem from three complementary perspectives: \textit{topological structure}, \textit{cross-domain dependencies}, and \textit{temporal evolution}. This multi-view analysis reveals distinct development patterns. General domains primarily expand horizontally, forming a wide and shallow structure (average depth 1.05) that exhibits signs of saturation. In contrast, math evolves vertically (average depth 2.92), driven by intensive reuse of core anchors to support deep recursive refinement. Cross-domain analysis further identifies code as a critical bridge between general and math, while highlighting the severe scarcity of specialized science data (only 44 nodes), which necessitates heavy reliance on upstream resources from other domains.

This explicit lineage transparency enables concrete diagnoses of structural issues. Our analysis shows that 17 of the 83 examined datasets exhibit redundancy rates exceeding 1\%, with \texttt{open-instruct-v1}~\cite{hakurei2023openinstructv1} reaching 46.48\% due to the inclusion of its own superset. Moreover, we uncover widespread benchmark contamination propagation: 19 datasets demonstrate varying degrees of leakage across benchmarks such as \texttt{Omni-Math}~\cite{gao2024omni}, \texttt{TheoremQA}~\cite{chen2023theoremqa}, \texttt{LiveCodeBench}~\cite{jain2024livecodebench}, \texttt{TruthfulQA}~\cite{lin2022truthfulqameasuringmodelsmimic}, and \texttt{SciBench}~\cite{wang2024scibenchevaluatingcollegelevelscientific}.
A notable example is \texttt{Caco-1.3M}~\cite{lin2025scalingcodeassistedchainofthoughtsinstructions}, which implicitly inherits 37.95\% of \texttt{Omni-Math} samples from contaminated upstream sources despite not explicitly including the benchmark itself. Unlike conventional sample-based scanning, our lineage-based framework exposes these latent structural intersections and enables efficient tracing of contamination sources along inheritance paths.

Beyond diagnosis, we demonstrate the practical value of data lineage analysis by using it to guide the construction of a \textit{lineage-aware diversity-oriented dataset} via provenance-based sampling, which anchors selection at upstream root sources to explicitly counteract redundancy induced by derivative reuse.
Furthermore, we include a discussion on how lineage-centric analysis enables a shift from sample-level comparison to topological reasoning over dataset evolution, offering advantages in matching efficiency, robustness to semantic drift, discovery of evolutionary patterns, and long-term ecosystem scalability.

Our contributions are summarized as follows:
(1) We introduce the concept of data lineage and propose a  multi-agent framework to reconstruct the evolutionary dependencies of post-training datasets.
(2) We analyze the ecosystem to characterize domain-specific evolution and reveal structural issues, specifically quantifying data redundancy and tracing the propagation of benchmark contamination.
(3) We propose a lineage-aware curation strategy to maximize query-level semantic diversity, achieving superior diversity metrics compared to datasets across varying scales.

\section{Related Work}

\subsection{Post-Training Data Construction}
Post-training data acquisition has evolved from the early aggregation of real-world annotations~\cite{longpre2023flancollectiondesigningdata,hendrycksmath2021,cobbe2021trainingverifierssolvemath,raffel2020exploring,together2023redpajama} to a multidimensional synthesis paradigm. Dominant strategies now include \textit{semantic evolution}~\cite{xu2024wizardlm,luo2024mmevolempoweringmultimodallarge,pei2025scalediff} for complexity enhancement, \textit{knowledge distillation}~\cite{mitra2024orcamath,tian2025correctanswersequaldistillation,guha2025openthoughtsdatarecipesreasoning} leveraging teacher CoT traces, and \textit{structured  fusion}~\cite{pei2025mathfusion,pan2025reststresstestinglarge} for composite reasoning for distribution refinement, alongside multimodal augmentations~\cite{shen2025evolvedgrpo,Yu_2025_CVPR}. Consequently, data constructed entirely ``from scratch'' has become rare~\cite{xu2024magpie,li2025cipherbank}. This widespread repurposing yields deeply nested dependencies, but their evolutionary pathways are seldom tracked.

\subsection{Data Analysis Paradigms and the Evolution of Sourcing}
In response to this entangled landscape, analysis tools have evolved from early documentation initiatives~\cite{knowyourdata,data-measurements} to dataset quality evaluation and policy/compliance checks. Current approaches range from quality-based filtering~\cite{liu2024makesgooddataalignment,chen2024alpagasustrainingbetteralpaca,lu2023instaginstructiontagginganalyzing} and cross-domain mixing analysis~\cite{li2025domainhelpothersdatacentric} to large-scale corpus profiling and licensing audits~\cite{elazar2024whatsbigdata,longpre2023dataprovenanceinitiativelarge}. 
Although the concept of "sourcing" has been effectively operationalized to trace model architectural history~\cite{zhao2025surveylargelanguagemodels} or attribute specific model behaviors to individual training instances~\cite{akyurek-etal-2022-towards,guu2023simfluencemodelinginfluenceindividual,pang2025largelanguagemodelsourcing}, these methods typically focus on the internal analysis of isolated datasets or specific samples, leaving the evolutionary relationships between datasets in the data ecosystem largely unexplored.

\section{Data Lineage}
\paragraph{Preliminary.}
We introduce a systematic framework for the automated tracing of data lineage, which is formally defined as a directed graph $\mathcal{G} = (\mathcal{V}, \mathcal{E})$, where each node in $\mathcal{V}$ corresponds to a post-training dataset. 
The nodes are categorized into two types: \textit{internal node}, which is the dataset with identifiable upstream sources that enable recursive tracing; and \textit{leaf node}, which denotes a terminal dataset that lacks such prerequisites and define the boundaries of automated exploration. Directed edges in $\mathcal{E}$ represent inheritance dependencies, where an edge $(v_i, v_j) \in \mathcal{E}$ indicates that the upstream dataset $v_i$ contributes to the construction of the dataset $v_j$.

\subsection{Challenges in Lineage Tracing}
Tracing lineage at scale is non-trivial due to the informal and heterogeneous nature of dataset documentation. Provenance information is scattered across various sources, such as academic papers, repository READMEs, and technical blogs, and is rarely expressed in a standardized format. Moreover, the dependency structure is often extensive and deeply nested: a single dataset may cite numerous upstream sources, and recursively expanding these references risks a combinatorial explosion in the search space. To address these challenges, we design a multi-agent collaborative framework that coordinates multi-source evidence fusion and semantic reasoning to extract structured lineage from noisy, incomplete documentation.

\begin{figure*}[t]
    \centering
    \includegraphics[width=\linewidth, trim=1.6cm 5.5cm 2cm 2cm, clip]{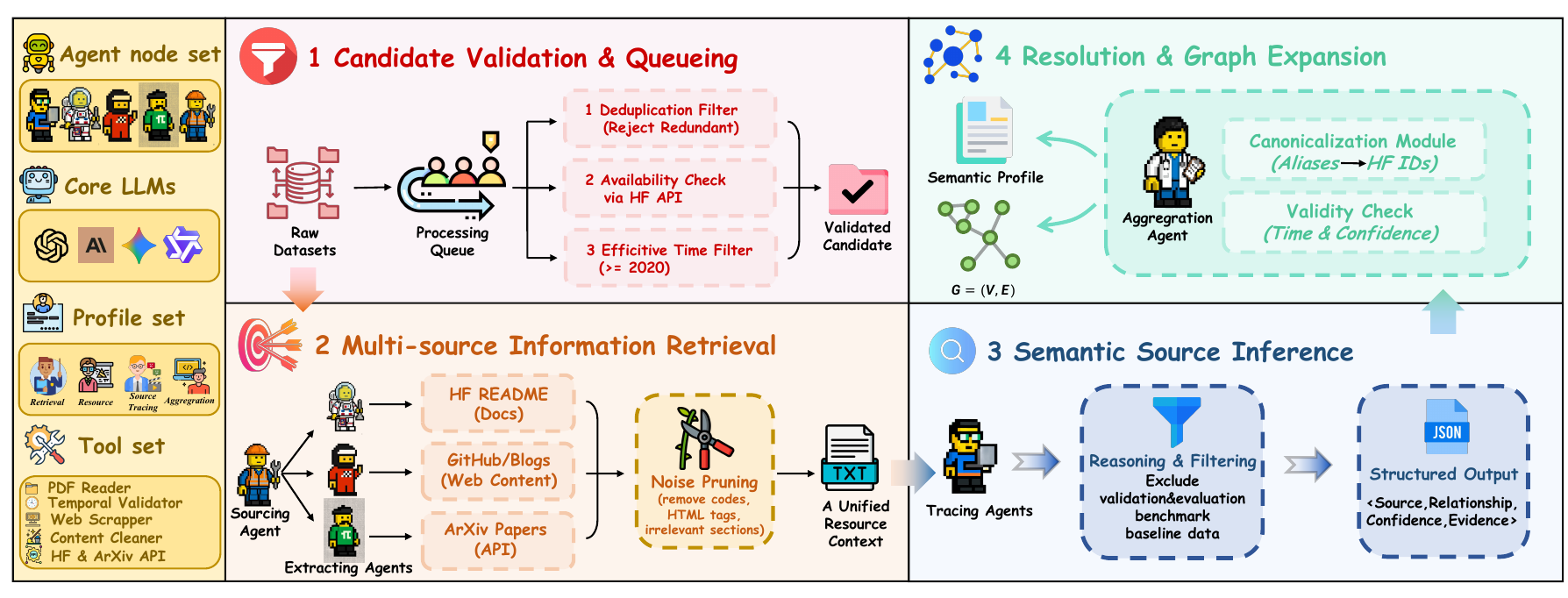}
    \caption{Overview of the multi-agent data lineage reconstruction framework. The system coordinates collaborative agents to extract provenance information from unstructured documentation, transforming isolated datasets into a comprehensive evolutionary graph.}
    \label{fig:framework}
    \vspace{-5mm}
\end{figure*}

\subsection{Multi-Agent Collaborative Framework}
As illustrated in Figure~\ref{fig:framework}, our framework operates via a target-to-source recursive pipeline managed through a centralized processing queue of pending datasets, designed to incrementally construct the lineage graph by tracing upstream dependencies. For each candidate dataset, the pipeline executes four sequential steps:

\paragraph{(1) Candidate Validation.}
We initialize the framework by enqueuing all candidate datasets into the centralized processing queue. For each candidate, we first filter out previously processed entries to prevent redundant computation and subsequently verify its availability via the HuggingFace API. To address potential latencies between research publication and repository upload, we determine the dataset's effective release time by cross-referencing its HuggingFace timestamp with the publication date of its associated paper, adopting the earlier of the two as the canonical release date. Finally, to align with the modern LLM era following GPT-3~\cite{brown2020languagemodelsfewshotlearners}, we restrict our analysis to datasets with an effective release time after 2020.

\paragraph{(2) Multi-source Information Retrieval.} For each validated candidate dataset, we issue a request to retrieve its HuggingFace README. We employ a \textit{sourcing agent} to parse the documentation and discover external resources, including GitHub repositories, technical blogs, and papers. Subsequently, we dispatch specialized \textit{extracting agents} to fetch the associated content. Specifically, agents retrieve web content for repositories and blogs, while querying the arXiv API for papers using titles or URLs. To enhance context quality and mitigate interference during subsequent lineage analysis, we apply a tailored filtering mechanism to eliminate structural noise such as metadata headers and code blocks in READMEs, HTML tags in blogs, and non-informative sections in papers. Finally, the curated materials are consolidated into a unified resource context.

\paragraph{(3) Semantic Source Inference and Extraction.}
Building on the consolidated resource context, we deploy a pool of \textit{tracing agents} operating in parallel to identify the source data utilized in constructing the candidate dataset. These agents are explicitly instructed to distinguish actual sources from incidental mentions, rigorously excluding entities such as evaluation benchmarks, comparison baselines, and non-integrated references. The extraction results are formalized as structured JSON records $\langle S, R, C, E \rangle$, where $S$ identifies the constituent source ancestor; $R$ categorizes the specific derivation relation (e.g., CoT distillation, question reformulation); $C$ quantifies identification confidence based on textual support strength and source credibility; and $E$ captures the supporting evidence. These records are aggregated to instantiate directed edges in the lineage graph.

\paragraph{(4) Aggregation, Resolution, and Recursive Expansion.}
The raw extraction records from parallel agents are first pooled and deduplicated by an \textit{aggregation agent} to eliminate redundancy across documentation sources. To address naming inconsistencies, the agent employs a retrieval-augmented resolution module that attempts to canonicalize informal aliases into unique HuggingFace IDs (i.e., org/name) via API verification and similarity reasoning. We subsequently enforce rigorous validity checks, pruning anachronistic edges where the source postdates the target and filtering out low-confidence hallucinations lacking verifiable evidence. Beyond structural lineage, the agent synthesizes a comprehensive semantic profile for the target dataset by integrating its inherent metadata with the composition of its upstream sources. This profile encapsulates key attributes including the dataset summary, capability domains, and construction methods. Finally, identified upstream sources are submitted to the centralized processing queue for subsequent recursive processing.

\paragraph{Graph Construction and Verification Strategy.}
Anchored within the ecosystem, our framework utilizes canonical org/name identifiers to execute a Depth-First Search (DFS) traversal over the dependency network. This recursion terminates at leaf nodes, identified by two convergence criteria: (1) foundational status lacking ancestors (upstream sources); or (2) release dates predating 2020. While preserved for completeness, these nodes halt further expansion. Crucially, to mitigate LLM hallucinations, we implement a confidence-aware expert verification protocol that automatically routes low-confidence extractions for manual review, ensuring the integrity of the final ecosystem map.

\section{Landscape Analysis}

\paragraph{Experimental Setup.}
We implement the lineage tracing framework using LangChain\footnote{\url{https://github.com/langchain-ai/langchain}} for workflow orchestration, leveraging GPT-5.1\footnote{specifically version \texttt{gpt-5.1-2025-11-13}; see \url{https://platform.openai.com/docs/models/gpt-5.1}.} and Gemini-2.5-flash~\cite{comanici2025gemini25pushingfrontier} as the underlying models for agent implementation. Notably, once deployed, the provenance system can trace newly encountered data in real time. To delineate the current landscape of the post-training ecosystem, we focus our analysis on the textual modality, as it represents the most prevalent data form. For dataset selection, we jointly consider HuggingFace downloads, repository likes, and citations, curating \textbf{83} high-impact textual datasets spanning four domains (general, math, code, and science) as seed roots for recursive traversal. Dataset details and framework configurations are provided in \Cref{sec:appendix_implementation}.

\begin{figure*}[t]
\centering
\includegraphics[width=\linewidth]{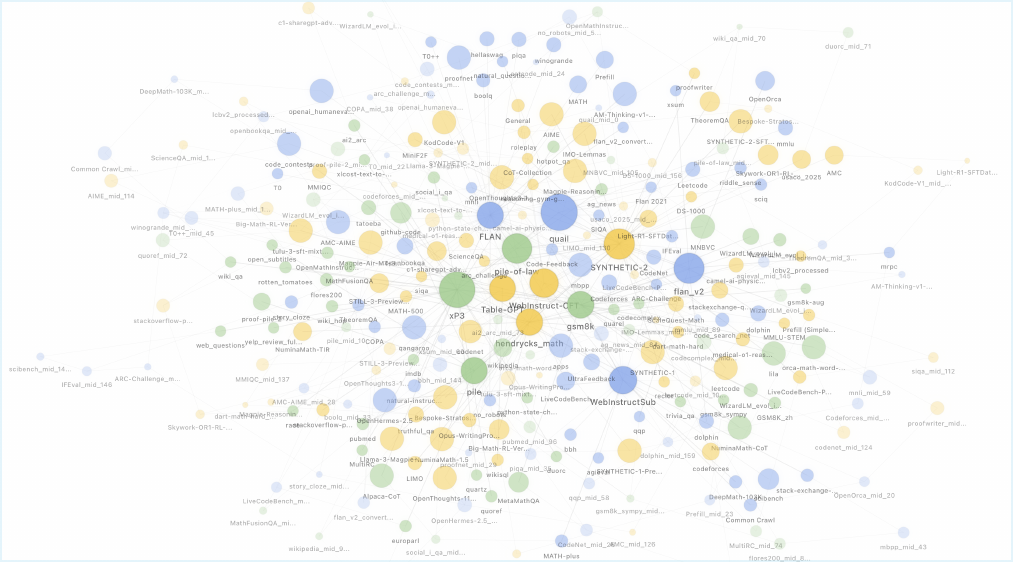}
\caption{Partial high-level overview of data lineage relationships, where node size reflects data download count, colors represent distinct data sub-networks, and darker shades indicate higher-degree, more important nodes.}
\label{fig:high}
\end{figure*}

\subsection{Global Topology and Evolutionary Trends}
\label{sec:topology}

\begin{table}[t]
\centering
\resizebox{0.63\columnwidth}{!}{
\begin{tabular}{lccccc}
\toprule
\textbf{Domain} & \textbf{Nodes} & \textbf{Depth} & \textbf{In-Deg.} & \textbf{Out-Deg.} & \textbf{Leaf \%} \\ 
\midrule
\textbf{Math}    & 99 & \textbf{2.92} & 3.30 & \textbf{1.54} & 38.38\% \\
\textbf{Code}    & 98 & 2.12 & 3.78 & 1.36 & 43.88\% \\
\textbf{General} & \textbf{285} & 1.05 & 2.51 & 1.29 & \textbf{68.42\%} \\
\textbf{Science} & 44 & 2.82 & \textbf{3.98} & 1.25 & 47.73\% \\
\bottomrule
\end{tabular}
}
\caption{Topological statistics by domain.}
\label{tab:domain_stats}
\end{table}

\begin{table}[t]
\centering
\resizebox{0.72\columnwidth}{!}{
\begin{tabular}{lcccc}
\toprule
\textbf{Source}$\to$\textbf{Target} & \textbf{Math} & \textbf{Code} & \textbf{General} & \textbf{Science} \\
\midrule
\textbf{Math}    & 147 (44.82\%) & 63 (17.80\%)  & 48 (9.66\%)   & 43 (23.24\%) \\
\textbf{Code}    & 67 (20.43\%)  & 118 (33.33\%) & 65 (13.08\%)  & 48 (25.95\%) \\
\textbf{General} & 73 (22.26\%)  & 137 (38.70\%) & 350 (70.42\%) & 64 (34.59\%) \\
\textbf{Science} & 41 (12.50\%)  & 36 (10.17\%)  & 34 (6.84\%)   & 30 (16.22\%) \\
\bottomrule
\end{tabular}
}
\caption{Cross-domain dependency matrix reordered by domain. Values denote counts (and column-wise percentages), indicating the proportion of a target domain's composition derived from each source domain.}
\label{tab:domain_dependency}
\end{table}

\begin{figure}[t]
\centering
\includegraphics[width=0.8\linewidth]{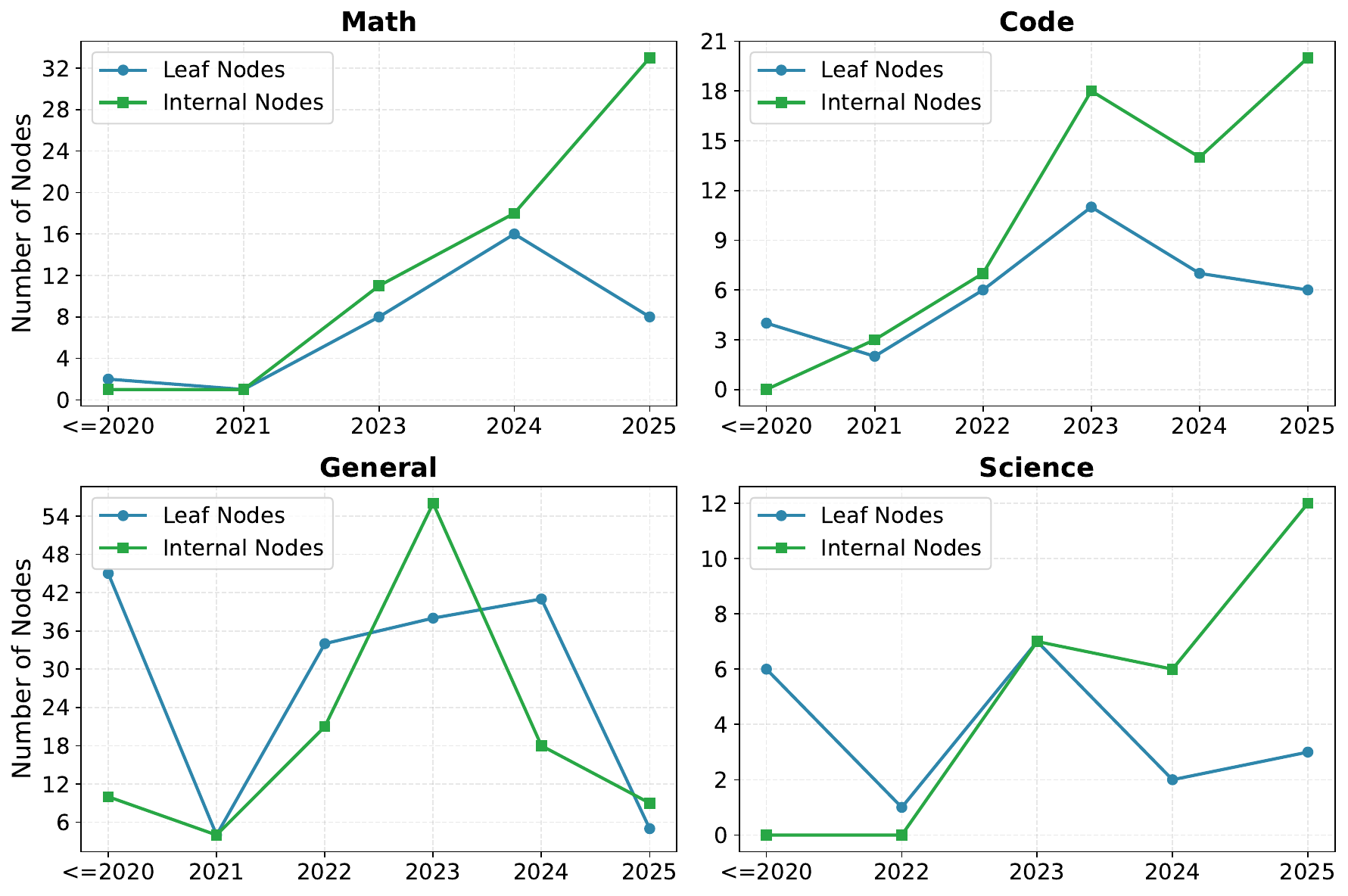}
\caption{Temporal distribution of data lineages by domain. The plots illustrate the number of datasets released each year, categorized by node type.}
\label{fig:alldomain}
\end{figure}

Starting from 83 high-impact seed datasets, our recursive lineage tracing reconstructed an expansive graph comprising 430 distinct datasets connected by 971 inheritance edges. By synthesizing graph topological metrics, cross-domain composition dependencies, and temporal evolutionary trajectories, we identify a fundamental structural transition characterized by divergent evolutionary strategies across domains.

\paragraph{Evolutionary Divergence: Broad Accumulation vs. Deep Refinement.}
The topology of the lineage graph reveals two differentiated evolutionary patterns across domains. 
(1) General-domain datasets exhibit a broad accumulation paradigm. Their shallow structures (leaf ratio 68.42\% and average depth 1.05, Table~\ref{tab:domain_stats}) reflect a strategy that prioritizes covering a wide range of topics by gathering information from many sources. This tendency to aggregate data is exemplified by massive collections like \texttt{FineWeb} ($d_{in}=111$, Table~\ref{tab:appendix_top_indegree_global}), while even the most reused dataset \texttt{FLAN} shows modest downstream influence ($d_{out}=7$, Table~\ref{tab:appendix_top_reused_domain}).
(2) Conversely, mathematics follows a deep refinement paradigm characterized by a vertically structured, recursive topology (average depth 2.92, average $d_{out}=1.54$). High-consensus anchors like \texttt{hendrycks\_math} ($d_{out}=19$) and \texttt{gsm8k} ($d_{out}=14$) serve as central roots supporting multiple generations of descendants. This is clearly observed in \texttt{NuminaMath} ($d_{out}=13$), which evolved from these foundational datasets and has now become a new cornerstone itself. Given this recursive structure, introducing novel, independent mathematical seeds is essential to complement synthetic augmentation and prevent saturation.

\paragraph{Cross-Domain Composition and Functional Specialization.}
The dependency matrix (Table~\ref{tab:domain_dependency}) reveals distinct functional roles within the ecosystem.
(1) Self-sourcing and independence. The general domain exhibits the highest independence, evidenced by a dominant 70.42\% self-sourcing rate. Math ranks second (44.82\%), demonstrating strong internal recycling while drawing on general inputs for linguistic context.
(2) Code as the operational link. Code operationalizes reasoning through a balanced profile: 38.70\% from general (capturing user intent) and 17.80\% from math (enhancing reasoning capabilities). This dual-sourcing positions Code datasets as a functional intermediary, translating abstract logic into verifiable, executable outputs.
(3) Science: Underdeveloped Status and High Dependence. With only 44 nodes, the Science domain remains underdeveloped. Its high external usage (average $d_{in}=3.98$) paired with low self-sourcing (16.22\%) indicates a heavy reliance on other domain resources due to the scarcity of native data. Consequently, future efforts should prioritize constructing specialized foundational datasets to bridge this gap.

\paragraph{Temporal Evolution: General Saturation vs. Specialized Growth.}
The temporal trajectories (Figure~\ref{fig:alldomain}) reveal a decisive shift in community prioritization.
(1) Saturation of broad acquisition. The general domain exhibits clear signs of saturation. New leaf node injection dropped sharply from 41 in 2024 to just 5 in 2025, marking the relative maturity of foundational natural language coverage and indicating that the phase of broad raw text acquisition has plateaued.
(2) Strategic prioritization of specialized reasoning. Conversely, focus has shifted toward specialized domains. Math witnessed a surge in intermediate nodes ($18 \to 33$ from 2024 to 2025), cementing its status as the core driver for logic enhancement. Similarly, science saw its intermediate output double ($6 \to 12$), signaling escalating attention to complex domain challenges. This divergence confirms a structural transition from broad knowledge accumulation to deep reasoning synthesis.

\definecolor{StaticColor}{RGB}{127, 60, 0}

\begin{figure*}[t]
    \centering
    \includegraphics[width=\linewidth, trim=0.5cm 2.5cm 0.5cm 2cm, clip]{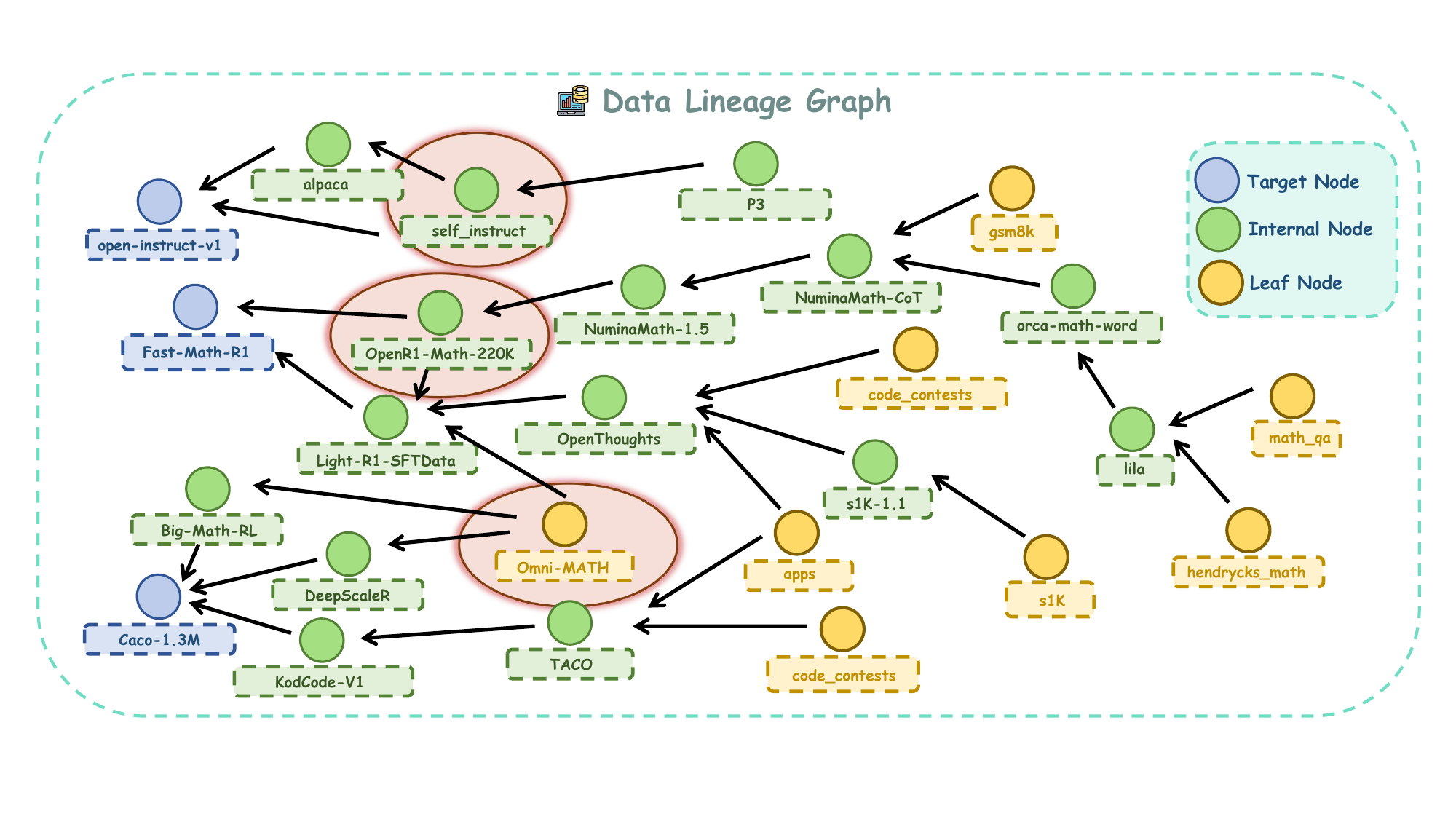}
    \caption{Schematic illustration of a small subgraph of the data lineage graph, showing the partial composition of three target datasets; downstream overlap nodes and benchmark contamination locations are highlighted in \textbf{\textcolor{StaticColor}{red}}.}
    \label{fig:graph}
\end{figure*}

\subsection{Analysis of Source Intersection}
\label{intersection}
As the community shifts toward a data-centric paradigm, dataset iteration has accelerated through the recursive integration of existing high-quality corpora. While this strategy allows developers to ``stand on the shoulders of giants''—minimizing cold-start costs while expanding scale—it introduces latent structural risks. Specifically, the lack of lineage transparency leads to unintended structural intersections, where seemingly distinct datasets unknowingly converge on identical upstream sources.

To expose latent structural intersections, our lineage tool constructs a dependency graph revealing upstream connections hidden within deep pathways. Using \texttt{Fast-Math-R1-SFT}~\cite{yoshihara2025practicaltwostagerecipemathematical} as a case study, our tool detected the concurrent incorporation of \texttt{OpenR1-Math-220k} and its superset, \texttt{Light-R1-SFTData}~\cite{lightr1proj}. A similar pattern recurs in \texttt{open-instruct-v1}, where \texttt{self\_instruct} is repeatedly included in its lineage (Figure~\ref{fig:graph}). To quantify this redundancy, we applied a strict metric based on exact (instruction, input, output) triplet matches. This analysis revealed unintended redundancy rates of 5.30\% and 46.48\%, respectively. Extending this scrutiny to the broader ecosystem, we identified similar nested patterns across multiple collections; the top 10 datasets exhibiting the highest structural redundancy are summarized in Table~\ref{tab:dataset_repetition_rates_sorted}.
\begin{table}[t]
\centering
\resizebox{0.5\columnwidth}{!}{
\begin{tabular}{lc}
\toprule
\textbf{Dataset Name} & \textbf{Rate(\%)} \\
\midrule
open-instruct-v1~\cite{hakurei2023openinstructv1} & 46.48 \\
opc-sft-stage2~\cite{Huang2024OpenCoderTO} & 27.96 \\
codeforces-cots~\cite{penedo2025codeforces} & 23.12 \\
Python-Code-23k-ShareGPT~\cite{bawase2023python} & 19.89 \\
CodeFeedback-Filtered-Instruction~\cite{opencodeinterpreter} & 8.00 \\
OpenMathInstruct-2~\cite{toshniwal2024openmath2} & 6.11 \\
Fast-Math-R1-SFT~\cite{yoshihara2025practicaltwostagerecipemathematical} & 5.30 \\
Open-Omega-Forge-1M~\cite{prithivMLmods2025openomega} & 4.33 \\
Light-R1-SFTData~\cite{lightr1proj} & 4.29 \\
OpenCodeReasoning~\cite{ahmadopencodereasoning} & 3.92 \\
\bottomrule
\end{tabular}
}
\caption{Analysis of source intersections across the top-10 datasets, ranked in descending order. Further details are provided in \Cref{sec:appendix_intersection}.}
\label{tab:dataset_repetition_rates_sorted}
\end{table}

In light of these findings, several recommendations are summarized (Rec.):
Rec.\ding{182} Prioritize orthogonal datasets via lineage analysis. For example, avoid adding OpenR1 if its superset Light-R1 is already selected, thereby maximizing diversity efficiency.
Rec.\ding{183} When multi-path ingestion is unavoidable, immediate deduplication is essential to eliminate structural redundancy caused by intersecting sources.

\subsection{Analysis of Benchmark Contamination}
\label{Contamination}
Data contamination, defined as the inadvertent inclusion of evaluation data into training corpora, undermines evaluation credibility. This issue creates a cascading effect where pollution in upstream datasets propagates downstream through the construction pipeline. Such propagation blurs the boundary between training and testing data, which creates a false sense of model capability.

Traditional decontamination methods, such as N-gram matching or semantic embedding retrieval~\cite{golchin2024timetravelllmstracing,li-etal-2024-open-source}, encounter significant limitations. Although precise, they require computationally expensive sample-wise scans and fail to map propagation across datasets. In contrast, data lineage offers a new, global perspective. It traces contamination diffusion along inheritance paths, which allows users to pinpoint upstream sources without full-scale content scanning.
\begin{figure}[t]
\centering
\includegraphics[width=0.75\linewidth]{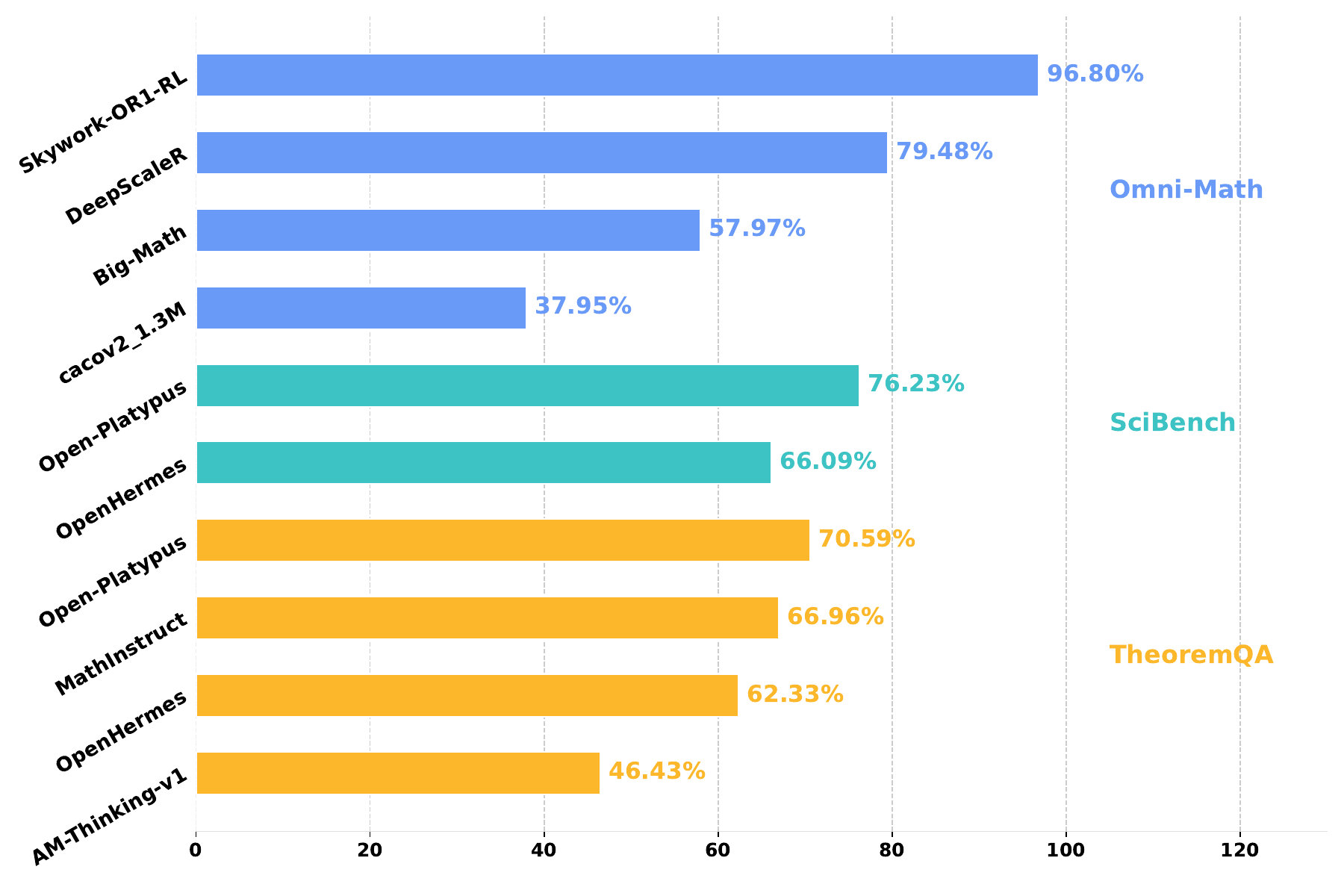}
\caption{Benchmark contamination analysis across various datasets. Additional contamination results are provided in \Cref{sec:appendix_contamination}.}
\label{fig:bench_contamination}
\end{figure}

Leveraging our lineage graph, we detect contamination across five benchmarks involving 19 datasets. As shown in Figure~\ref{fig:graph} and~\ref{fig:bench_contamination}, \texttt{DeepScaleR-Preview-Dataset} and \texttt{Big-Math-RL-Verified} directly ingest \texttt{omni-math}, which results in leakage rates of 79.48\% and 57.97\% respectively. Consequently, \texttt{Caco-1.3M} inadvertently inherits 37.95\% contamination by incorporating these datasets. 

We propose the following protocols for data curators (Rec.):
Rec.\ding{182} Exclude compromised benchmarks. If upstream sources contain benchmark data, strict exclusion of the corresponding evaluation sets is imperative to prevent performance inflation.
Rec.\ding{183} Pre-screen via lineage analysis. Before integrating external data, lineage tools audit upstream composition to intercept unintentional leakage at the source.

\section{Analysis and Discussion}
\label{sec:discussion}

The global data lineage graph supports both diagnosing issues such as contamination and guiding data curation. We discuss how topological signals can improve dataset construction, and summarize the benefits of lineage-centric analysis over sample-level methods.

\subsection{Lineage-Guided Data Construction}
\label{sec:diversity_optimization}

In post-training data construction, the instruction (query) serves as a stable semantic anchor. While response distillation is inherently constrained by teacher models, instructions often remain invariant across multiple rounds of lineage reuse. Consequently, recursive dataset derivation frequently induces implicit redundancy, narrowing the effective problem space~\cite{sandholm2024randomness}.

To address this, we propose \textit{provenance-based sampling}, a lineage-aware strategy designed to maximize instruction diversity. Rather than sampling from the entire corpus, we treat root node datasets ($d_{in}=0$) as upstream knowledge anchors. We prioritize these sources based on both their domain metadata and topological influence, quantified by out-degree. Following this selection, we apply MinHash~\cite{broder1997syntactic} for duplicate removal. This yields a lineage-aware dataset with 570K unique instructions.

\paragraph{Empirical Validation and Potential.}
We evaluate diversity using the Vendi Score~\cite{friedman2023vendiscorediversityevaluation} and Centroid Distance~\cite{suwanda2020analysis}. To ensure a rigorous evaluation, we benchmarked our approach against a diverse array of datasets ranging from 300K to 1.2M samples. We selected baselines that are widely recognized for their quality and coverage rather than domain-specific niches. As presented in Table~\ref{tab:diversity_comparison}, our strategy achieves a Vendi Score of \textbf{452.44} and a Centroid Distance of \textbf{0.6385}. These results demonstrate that our method delivers superior performance across the entire spectrum of data scales. Notably, our dataset outperforms \texttt{OpenHermes-2.5}, a strong baseline renowned in the community for its extensive topic coverage. Even more significantly, our approach substantially exceeds much larger collections such as \texttt{MegaScience} and \texttt{OpenThoughts3}. Despite these datasets containing more than double the number of samples compared to ours, they exhibit lower diversity scores. 

This empirical evidence shows that larger data volume does not automatically translate to higher semantic diversity scores. Our lineage-guided approach achieves superior diversity metrics at a smaller scale, demonstrating the effectiveness of provenance-based sampling in maximizing instruction coverage.

Notably, these results are obtained using only root nodes: we intentionally exclude internal evolutionary variants to isolate the effect of provenance disentanglement. That this root-only subset outperforms complex mixtures suggests two implications. 
\paragraph{(1) Efficiency:} Preserving original provenance can yield high diversity without exhaustive filtering over the derivative space. 
\paragraph{(2) High Ceiling:} Since refined internal nodes with richer rewrites and semantic variants are not yet included, there remains substantial headroom; incorporating these hubs may further improve dataset quality.

\begin{table}[t]
\centering
\resizebox{0.65\columnwidth}{!}{
\begin{tabular}{lccc}
\toprule
\multirow{2}{*}{\textbf{Dataset}} & \multirow{2}{*}{\textbf{Size}} & \multicolumn{2}{c}{\textbf{Diversity Metric}} \\
\cmidrule(lr){3-4}
& & \textbf{Vendi Score} $\uparrow$ & \textbf{Cent. Dist.} $\uparrow$ \\
\midrule
OmniThought-0528~\cite{cai2025reasoningomnithoughtlargecot} & 301K & 162.52 & 0.5140 \\
herculesv1~\cite{HerculesV1.0} & 463K & 397.33 & 0.6121 \\
OpenHermes-2.5~\cite{OpenHermes2.5} & 615K & \underline{437.76} & \underline{0.6271} \\
TextbookReasoning~\cite{fan2025megascience} & 651K & 283.75 & 0.5598 \\
MiroMind-M1-SFT~\cite{li2025miromindm1opensourceadvancementmathematical} & 719K & 108.89 & 0.4597 \\
tulu-3-sft-mixture~\cite{lambert2024tulu3} & 939K & 375.78 & 0.6169 \\
OpenThoughts3~\cite{guha2025openthoughtsdatarecipesreasoning} & 1.2M & 133.26 & 0.4970 \\
MegaScience~\cite{fan2025megascience} & 1.2M & 373.78 & 0.6150 \\
\midrule
\textbf{Ours (Provenance-based)} & \textbf{570K} & \textbf{452.44} & \textbf{0.6385} \\
\bottomrule
\end{tabular}
}
\caption{Diversity comparison using Vendi Score and Centroid Distance.}
\label{tab:diversity_comparison}
\end{table}

\subsection{Lineage vs. Sample-Level Analysis}
\label{sec:advantages}

Beyond dataset construction, the lineage graph provides a more efficient framework for data analysis compared to traditional sample-level methods. We highlight three key advantages.

\paragraph{Efficiency in Dataset Matching.}
Sample-level similarity checks require scanning and comparing large numbers of examples, which is costly at scale. Lineage analysis shifts the unit of comparison from samples to ancestry. By comparing the overlap of upstream sources and inheritance paths, we can quickly estimate whether two datasets are likely to share content and where the overlap comes from, without traversing millions of samples.

\paragraph{Discovery of Evolutionary Paradigms.}
The graph groups related datasets into families and makes their construction steps explicit. By tracing recurring parent-to-child transformations in successful lineages, we can summarize common build patterns, such as textbook sources, synthetic Q\&A generation, and CoT refinement. These patterns offer practical guidance on which sources and refinement steps often co-occur in high-impact datasets.

\paragraph{Robustness Against Semantic Drift.}
Content-based matching is fragile when data is rewritten, expanded, or reformatted, since surface similarity can disappear. Lineage analysis relies on dependency links and provenance records rather than text overlap, so it can still connect an evolved dataset to its origins even after substantial edits. This preserves traceability and supports reliable auditing of redundancy or contamination.

\section{Conclusion}
In this paper, we introduce a multi-agent collaborative framework to reconstruct a large-scale data lineage graph for post-training datasets. Our analysis uncovers structural redundancy and traces the propagation of benchmark contamination across the ecosystem. Furthermore, we discussed the application of the lineage graph in data construction, focusing on diversity, and outlined potential directions for future research.

\section*{Limitations}
Our framework faces two primary limitations. First, relying on LLMs entails inherent hallucination risks, necessitating human verification for low-confidence extractions to ensure graph reliability. Second, our lineage reconstruction is strictly bound by documentary transparency; the system cannot recover dependencies if dataset creators fail to report or intentionally conceal upstream sources in their technical documentation.

\clearpage
\newpage
\bibliographystyle{plainnat}
\setcitestyle{numbers}
\bibliography{paper.bib}

\clearpage
\newpage
\beginappendix

\section{Implementation Details of Automatic Provenance Framework}
\label{sec:appendix_implementation}

In this section, we provide a detailed overview of the pipeline used to construct the data lineage graph. We discuss the specific models employed for each agent within our framework and list the seed datasets selected for the initial analysis.

\subsection{Agent Model Configuration}
To balance accuracy, efficiency, and cost, we assigned different large language models to specific agents based on their performance characteristics.

\paragraph{Sourcing Agent.}
This agent is responsible for accurately identifying entry points for data information from repository README files. To minimize hallucination risks and ensure precise extraction, we utilized \texttt{GPT-5.1}. Its strong instruction-following capability ensures that the initial retrieval of metadata is reliable.

\paragraph{Extracting Agent.}
This agent visits the identified sources and summarizes the information into a specified format. Given the high volume of text processing required, we selected \texttt{Gemini-2.5-Flash}. Its high processing speed allows for efficient large-scale information scanning and summarization without improved latency.

\paragraph{Tracing Agent.}
After information summary, this agent extracts the specific source-target relationships. Similar to the Sourcing Agent, precision is critical here to avoid false lineages. Therefore, we again employed \texttt{GPT-5.1} to leverage its low hallucination rate for accurate relationship extraction.

\paragraph{Aggregation Agent.}
This agent handles the standardization of dataset names and the merging of sources. We utilized \texttt{Gemini-2.5-Pro} for this task. Its strong reasoning capabilities and support for web retrieval allow it to resolve ambiguous dataset names by verifying them against online resources. It effectively consolidates dispersed information into unified nodes.

\subsection{Seed Data Selection}
We curated a set of 83 seed datasets to serve as the starting point for our lineage analysis. These datasets were selected based on three criteria: download volume, community engagement (likes or stars), and the reported performance of downstream models trained on them. All selected datasets are verifiable and retrievable on Hugging Face. The complete list is provided in Table~\ref{tab:seed_datasets}.

\subsection{Relation Types in Data Lineage}
\label{app:relation_types}

The relation type $R$ in our lineage graph characterizes the specific derivation methods between datasets. We identify five primary categories:

\begin{itemize}
    \item \textbf{Semantic Evolution}: Reformulating or enhancing the original question while allowing controlled semantic variation.
    \item \textbf{CoT Distillation}: Keeping the question unchanged while using a stronger teacher model to generate long CoT responses.
    \item \textbf{Synthetic Generation}: Using upstream data as seeds to generate new question-answer pairs through LLM generalization.
    \item \textbf{Structured Fusion}: Concatenating or combining data from multiple sources for composite reasoning.
    \item \textbf{Direct Inclusion/Subset}: Including upstream data as-is as part of the new dataset.
\end{itemize}

\paragraph{Role in graph construction.}
The tracing agent infers relation type $R$ by analyzing method descriptions in documentation (papers, READMEs, etc.) and stores it as an edge attribute in the graph. It should be noted that some datasets have vague descriptions (e.g., only mentioning "based on XX" without specifying the processing method). In such cases, the agent makes the best inference based on contextual semantics or labels it as the default "Direct Inclusion" relation.

Storing $R$ as an edge attribute serves two key purposes: (1) it enhances graph semantic richness by enabling the graph to answer not only "dataset A is derived from dataset B" but also "through what method," and (2) it supports downstream applications such as data deduplication, contamination tracking, and quality assessment by providing fine-grained signals about derivation patterns.

\begin{table}[t]
\centering
\caption{Full list of the 83 seed datasets used for lineage analysis. The datasets are sourced from HuggingFace and cover math, code, general and science domains.}
\label{tab:seed_datasets}
\scriptsize 
\renewcommand{\arraystretch}{1.15} 
\setlength{\tabcolsep}{4pt}
\begin{tabular}{p{0.32\linewidth} p{0.32\linewidth} p{0.32\linewidth}}
\toprule
\multicolumn{3}{c}{\textbf{Seed Datasets List}} \\
\midrule
\texttt{m-a-p/CodeFeedback-Filtered-Instruction} & \texttt{garage-bAInd/Open-Platypus} & \texttt{amphora/QwQ-LongCoT-130K} \\
\texttt{open-thoughts/OpenThoughts3-1.2M} & \texttt{WizardLMTeam/WizardLM\_evol\_instruct\_V2\_196k} & \texttt{ajibawa-2023/Code-74k-ShareGPT} \\
\texttt{a-m-team/AM-Thinking-v1-Distilled} & \texttt{gretelai/gretel-text-to-python} & \texttt{QuixiAI/dolphin} \\
\texttt{microsoft/rStar-Coder} & \texttt{TokenBender/code\_instructions\_122k} & \texttt{EricLu/SCP-116K} \\
\texttt{OpenCoder-LLM/opc-sft-stage2} & \texttt{sequelbox/Raiden-DeepSeek-R1} & \texttt{GAIR/o1-journey} \\
\texttt{Locutusque/hercules-v1.0} & \texttt{Magpie-Align/Magpie-Reasoning-V2-250K} & \texttt{miromind-ai/MiroMind-M1-SFT-719K} \\
\texttt{theblackcat102/evol-codealpaca-v1} & \texttt{PrimeIntellect/SYNTHETIC-2-SFT-verified} & \texttt{zwhe99/DeepMath-103K} \\
\texttt{ajibawa-2023/Code-290k-ShareGPT} & \texttt{MegaScience/TextbookReasoning} & \texttt{open-r1/OpenR1-Math-220k} \\
\texttt{KodCode/KodCode-V1} & \texttt{qihoo360/Light-R1-SFTData} & \texttt{dyyyyyyyy/ScaleQuest-Math} \\
\texttt{m-a-p/Code-Feedback} & \texttt{Magpie-Align/Magpie-Reasoning-V2-250K} & \texttt{AI-MO/NuminaMath-1.5} \\
\texttt{amphora/QwQ-LongCoT-130K-2} & \texttt{Magpie-Align/Magpie-Reasoning-V1-150K} & \texttt{AI-MO/NuminaMath-CoT} \\
\texttt{allenai/tulu-3-sft-mixture} & \texttt{O1-OPEN/OpenO1-SFT} & \texttt{PawanKrd/math-gpt-4o-200k} \\
\texttt{OpenCoder-LLM/opc-sft-stage1} & \texttt{RabotniKuma/Fast-Math-R1-SFT} & \texttt{bespokelabs/Bespoke-Stratos-17k} \\
\texttt{microsoft/EpiCoder-func-380k} & \texttt{SkunkworksAI/reasoning-0.01} & \texttt{hkust-nlp/dart-math-hard} \\
\texttt{MegaScience/MegaScience} & \texttt{tatsu-lab/alpaca} & \texttt{TIGER-Lab/WebInstruct-CFT} \\
\texttt{microsoft/orca-agentinstruct-1M-v1} & \texttt{vicgalle/alpaca-gpt4} & \texttt{rubenroy/GammaCorpus-CoT-Math-170k} \\
\texttt{alibaba-pai/OmniThought-0528} & \texttt{allenai/tulu-3-sft-personas-math} & \texttt{allenai/tulu-3-sft-personas-algebra} \\
\texttt{ajibawa-2023/Python-Code-23k-ShareGPT} & \texttt{QizhiPei/MathFusionQA} & \texttt{TIGER-Lab/MATH-plus} \\
\texttt{likaixin/InstructCoder} & \texttt{whynlp/gsm8k-aug} & \texttt{GAIR/LIMO} \\
\texttt{Mxode/Magpie-Pro-10K-GPT4o-mini} & \texttt{nvidia/OpenMathInstruct-2} & \texttt{microsoft/orca-math-word-problems-200k} \\
\texttt{efficientscaling/Z1-Code-Reasoning-107K} & \texttt{ajibawa-2023/Maths-College} & \texttt{ServiceNow-AI/R1-Distill-SFT} \\
\texttt{teknium/OpenHermes-2.5} & \texttt{agentica-org/DeepScaleR-Preview-Dataset} & \texttt{databricks/databricks-dolly-15k} \\
\texttt{ise-uiuc/Magicoder-OSS-Instruct-75K} & \texttt{gretelai/synthetic\_text\_to\_sql} & \texttt{open-r1/codeforces-cots} \\
\texttt{nickrosh/Evol-Instruct-Code-80k-v1} & \texttt{QizhiPei/ScaleDiff-Math} & \texttt{LHL3341/Caco-1.3M} \\
\texttt{WizardLMTeam/WizardLM\_evol\_instruct\_70k} & \texttt{gretelai/gretel-math-gsm8k-v1} & \texttt{prithivMLmods/Open-Omega-Forge-1M} \\
\texttt{open-thoughts/OpenThoughts-114k} & \texttt{OpenCoder-LLM/opc-sft-stage1} & \texttt{hakurei/open-instruct-v1} \\
\texttt{GAIR/lima} & \texttt{allenai/omega-explorative} & \texttt{openbmb/UltraInteract\_sft} \\
\texttt{allenai/tulu-3-sft-personas-code} & \texttt{bigcode/self-oss-instruct-sc2-exec-filter-50k} & \\
\bottomrule
\end{tabular}
\end{table}

\section{Topological Structure of the Lineage Graph}
\label{sec:appendix_topology}

Based on the extensive lineage graph constructed, we provide a detailed statistical analysis of the ecosystem's topological properties. We focus on three key dimensions to characterize the roles and evolutionary patterns of different datasets: reuse rate (measured by out-degree), information aggregation (measured by in-degree), and evolutionary depth. The specific statistics for these dimensions are presented in Table~\ref{tab:appendix_top_reused_domain}, Table~\ref{tab:appendix_top_indegree_global}, and Table~\ref{tab:appendix_top_depth}, respectively.

\begin{table}[h]
\centering
\small
\caption{Top 5 most reused datasets by domain (ranked by out-degree).}
\label{tab:appendix_top_reused_domain}

\begin{tabular*}{\textwidth}{@{\extracolsep{\fill}}cllcc}
\toprule
\textbf{Domain} & \textbf{Rank} & \textbf{Dataset Name} & \textbf{In-Degree} & \textbf{Release Date} \\
\midrule
\multirow{5}{*}{\textbf{Math}} 
 & 1 & \texttt{EleutherAI/hendrycks\_math} & 19 & 2021-03 \\
 & 2 & \texttt{openai/gsm8k} & 14 & 2021-10 \\
 & 3 & \texttt{AI-MO/NuminaMath-CoT} & 13 & 2024-07 \\
 & 4 & \texttt{open-r1/OpenR1-Math-220k} & 6 & 2025-02 \\
 & 5 & \texttt{meta-math/MetaMathQA} & 6 & 2023-09 \\
\midrule
\multirow{5}{*}{\textbf{Code}} 
 & 1 & \texttt{BAAI/TACO} & 11 & 2023-12 \\
 & 2 & \texttt{codeparrot/apps} & 10 & 2021-05 \\
 & 3 & \texttt{deepmind/code\_contests} & 9 & 2021-05 \\
 & 4 & \texttt{sahil2801/CodeAlpaca-20k} & 6 & 2023-03 \\
 & 5 & \texttt{ise-uiuc/Magicoder-Evol-Instruct-110K} & 5 & 2023-12 \\
\midrule
\multirow{5}{*}{\textbf{General}} 
 & 1 & \texttt{Open-Orca/FLAN} & 7 & 2021-09 \\
 & 2 & \texttt{tatsu-lab/alpaca} & 6 & 2023-03 \\
 & 3 & \texttt{anon8231489123/ShareGPT\_Vicuna\_unfiltered} & 5 & 2023-04 \\
 & 4 & \texttt{teknium/GPTeacher-General-Instruct} & 4 & 2023-04 \\
 & 5 & \texttt{wikimedia/wikipedia} & 4 & 2022-03 \\
\midrule
\multirow{5}{*}{\textbf{Science}} 
 & 1 & \texttt{open-thoughts/OpenThoughts-114k} & 3 & 2025-01 \\
 & 2 & \texttt{camel-ai/chemistry} & 3 & 2023-03 \\
 & 3 & \texttt{camel-ai/physics} & 3 & 2023-03 \\
 & 4 & \texttt{camel-ai/biology} & 3 & 2023-03 \\
 & 5 & \texttt{nvidia/Llama-Nemotron-Post-Training-Dataset} & 3 & 2025-03 \\
\bottomrule
\end{tabular*}
\end{table}

\begin{table}[h]
\centering
\small
\caption{Top 10 datasets with highest global in-degree.}
\label{tab:appendix_top_indegree_global}
\begin{tabular*}{\textwidth}{@{\extracolsep{\fill}}cllcc}
\toprule
\textbf{Rank} & \textbf{Dataset Name} & \textbf{Domain} & \textbf{In-Degree} & \textbf{Release Date} \\
\midrule
1 & \texttt{HuggingFaceFW/fineweb} & General & 111 & 2021-09 \\
2 & \texttt{bigscience/xP3} & Code, General & 79 & 2022-10 \\
3 & \texttt{CohereLabs/aya\_collection} & General & 70 & 2024-01 \\
4 & \texttt{bigscience/P3} & General & 54 & 2021-10 \\
5 & \texttt{allenai/lila} & Math & 20 & 2023-02 \\
6 & \texttt{izumi-lab/llm-japanese-dataset-vanilla} & General & 20 & 2023-05 \\
7 & \texttt{a-m-team/AM-Thinking-v1-Distilled} & Math, Code, Science, General & 19 & 2025-05 \\
8 & \texttt{teknium/OpenHermes-2.5} & Math, Code, Science, General & 19 & 2023-11 \\
9 & \texttt{izumi-lab/llm-japanese-dataset} & General & 19 & 2023-04 \\
10 & \texttt{allenai/tulu-3-sft-mixture} & Math, Code, Science, General & 18 & 2024-11 \\
\bottomrule
\end{tabular*}
\end{table}

\begin{table}[h]
\centering
\small
\caption{Top 5 datasets with highest evolutionary depth.}
\label{tab:appendix_top_depth}
\begin{tabular*}{\textwidth}{@{\extracolsep{\fill}}cllcc}
\toprule
\textbf{Rank} & \textbf{Dataset Name} & \textbf{Domain} & \textbf{Depth} & \textbf{Release Date} \\
\midrule
1 & \texttt{alibaba-pai/OmniThought} & Code, General, Math & 9 & 2025-05 \\
2 & \texttt{allenai/tulu-3-sft-mixture} & Math, Code, Science, General & 8 & 2024-11 \\
3 & \texttt{zwhe99/DeepMath-103K} & Math & 8 & 2025-04 \\
4 & \texttt{CohereForAI/aya\_dataset} & General & 7 & 2024-01 \\
5 & \texttt{open-thoughts/OpenThoughts2-1M} & Science, Code, General, Math & 7 & 2025-04 \\
\bottomrule
\end{tabular*}
\end{table}

\section{Source Intersection Details}
\label{sec:appendix_intersection}

As discussed in Section~\ref{intersection}, we conducted a rigorous intersection analysis using a strict matching protocol. We calculated hash values based on the complete \texttt{(instruction, input, output)} triplets to identify exact duplicates. Even under this strict criterion, we detected significant redundancy across multiple datasets.

Table~\ref{tab:selected_lineage} details the specific upstream intersection paths identified in our analysis. These paths reveal how certain datasets inadvertently incorporate large portions of upstream sources. We recommend that data curators adopt this verification method to detect hidden upstream intersections. This practice is essential for preventing the redundant selection of identical data sources when constructing a new training pool.

\section{Data Contamination Details}
\label{sec:appendix_contamination}

Expanding on the analysis in Section~\ref{Contamination}, this section details the downstream propagation of data contamination. After obtaining lineage relationships, we perform strict exact matching of benchmark samples \texttt{(instruction, input)} against downstream training data to measure actual contamination rates. We broaden the scope from the main text to include comprehensive statistics for five benchmarks, incorporating LiveCodeBench and TruthfulQA.

Special attention is required for LiveCodeBench due to its chronological update mechanism. Our analysis reveals that datasets such as \texttt{a-m-team/AM-Thinking-v1-Distilled} and \texttt{agentica-org/DeepCoder- Preview-Dataset} inadvertently incorporated test samples from LiveCodeBench v5\footnote{\url{https://livecodebench.github.io/leaderboard_v5.html}}. This exposes the critical risks associated with temporal benchmarks when strict version control and temporal cutoffs are neglected.

Table~\ref{tab:data_contamination} provides granular contamination statistics, reporting both the exact count and percentage of leaked samples. Additionally, Table~\ref{tab:data_lineage} traces the specific lineage pathways through which contamination enters the ecosystem. These findings empirically confirm that contamination in upstream sources inevitably cascades into downstream derivatives.

Consequently, we strongly advocate for the strict exclusion of benchmark samples from training corpora. Even if a model is not intended for evaluation on a specific benchmark, retaining these samples contaminates the shared data lineage, compromising future research. We emphasize that rigorous decontamination protocols are prerequisite to ensuring the validity of generalization assessments and preventing inflated evaluation metrics.

\begin{table}[t]
\centering
\small
\renewcommand{\arraystretch}{1.5} 
\renewcommand{\tabularxcolumn}[1]{m{#1}}

\newcommand{\code}[1]{\texttt{#1}}
\newcommand{\NextPath}{\par\vspace{3pt}\noindent}

\caption{Full lineage paths: tracing the complete evolution from target to source (selected datasets).}
\label{tab:selected_lineage}

\begin{tabularx}{\textwidth}{
    >{\centering\arraybackslash\small}m{3.5cm} 
    >{\raggedright\arraybackslash\small}X 
    >{\centering\arraybackslash\small}m{3cm}
}
\toprule
\textbf{Target Dataset} & \textbf{Complete Evolutionary Paths} & \textbf{Intersection Point} \\
\midrule

\textbf{\code{CodeFeedback-\newline Filtered-Instruction}} & 
\noindent\textbf{Path 1:} \code{m-a-p\slb CodeFeedback-Filtered-Instruction} \arr \code{ise-uiuc\slb Magicoder-Evol-Instruct-110K} \arr \code{theblackcat102\slb evol-codealpaca-v1} \arr \code{HuggingFaceH4\slb CodeAlpaca\_20K} \arr \code{sahil2801\slb CodeAlpaca-20k} 
\NextPath
\textbf{Path 2:} \code{m-a-p\slb CodeFeedback-Filtered-Instruction} \arr \code{nickrosh\slb Evol-Instruct-Code-80k-v1} \arr \code{sahil2801\slb CodeAlpaca-20k}
& 
\code{sahil2801\slb CodeAlpaca-20k} \\
\midrule

\textbf{\code{Fast-Math-R1-SFT}} & 
\noindent\textbf{Path 1:} \code{RabotniKuma\slb Fast-Math-R1-SFT} \arr \code{open-r1\slb OpenR1-Math-220k} 
\NextPath
\textbf{Path 2:} \code{RabotniKuma\slb Fast-Math-R1-SFT} \arr \code{qihoo360\slb Light-R1-SFTData} \arr \code{open-r1\slb OpenR1-Math-220k} 
& 
\code{open-r1\slb OpenR1-Math-220k} \\
\midrule

\textbf{\code{Open-Omega-\newline Forge-1M}} & 
\noindent\textbf{Path 1:} \code{prithivMLmods\slb Open-Omega-Forge-1M} \arr \code{nvidia\slb OpenCodeReasoning} \arr \code{codeparrot\slb apps} 
\NextPath
\textbf{Path 2:} \code{prithivMLmods\slb Open-Omega-Forge-1M} \arr \code{nvidia\slb OpenMathReasoning} \arr \code{nvidia\slb Llama-Nemotron-Post-Training-Dataset} \arr \code{codeparrot\slb apps} 
& 
\code{codeparrot\slb apps} \\
\midrule

\multirow{5}{=}{\centering\textbf{\code{Light-R1-SFTData}}} & 
\noindent\textbf{Path 1:} \code{qihoo360\slb Light-R1-SFTData} \arr \code{open-r1\slb OpenR1-Math-220k} \arr \code{AI-MO\slb NuminaMath-CoT} 
\NextPath
\textbf{Path 2:} \code{qihoo360\slb Light-R1-SFTData} \arr \code{open-thoughts\slb OpenThoughts-114k} \arr \code{AI-MO\slb NuminaMath-CoT} 
& 
\code{AI-MO\slb NuminaMath-CoT} \\
\cmidrule(lr){2-3}
& 
\noindent\textbf{Path 3:} \code{qihoo360\slb Light-R1-SFTData} \arr \code{nvidia\slb OpenMathInstruct-2} \arr \code{EleutherAI\slb hendrycks\_math} 
\NextPath
\textbf{Path 4:} \code{qihoo360\slb Light-R1-SFTData} \arr \code{GAIR\slb LIMO} \arr \code{EleutherAI\slb hendrycks\_math} 
& 
\code{EleutherAI\slb hendrycks\_math} \\
\midrule

\textbf{\code{OpenCodeReasoning}} & 
\noindent\textbf{Path 1:} \code{nvidia\slb OpenCodeReasoning} \arr \code{BAAI\slb TACO} \arr \code{codeparrot\slb apps} 
\NextPath
\textbf{Path 2:} \code{nvidia\slb OpenCodeReasoning} \arr \code{codeparrot\slb apps} 
& 
\code{codeparrot\slb apps} \\
\midrule

\textbf{\code{open-instruct-v1}} & 
\noindent\textbf{Path 1:} \code{hakurei\slb open-instruct-v1} \arr \code{tatsu-lab\slb alpaca} \arr \code{yizhongw\slb self\_instruct} 
\NextPath
\textbf{Path 2:} \code{hakurei\slb open-instruct-v1} \arr \code{yizhongw\slb self\_instruct} 
& 
\code{yizhongw\slb self\_instruct} \\

\bottomrule
\end{tabularx}
\end{table}

\section{Diversity Optimization via Root Nodes}
\label{sec:appendix_applications}

\paragraph{Provenance-based sampling workflow.}
In our provenance-based sampling strategy, we leverage leaf nodes ($d_{in}=0$) as \textbf{upstream knowledge anchors} to construct a diversity-oriented dataset. By anchoring sampling on leaf nodes, we reduce structural redundancy at the source, as these nodes represent independent original knowledge sources. The complete workflow proceeds as follows:

\begin{itemize}
    \item \textbf{Leaf node selection}: Filter all $d_{in}=0$ leaf nodes from the lineage graph (212 datasets) and rank by $d_{out}$, excluding nodes with zero downstream usage.
    \item \textbf{Domain filtering}: Retain domains commonly targeted in post-training (math, code, science, reasoning, etc.), filter out niche domains, and exclude non-QA format data (e.g., Common Crawl, Wikipedia), resulting in 31 datasets.
    \item \textbf{Format unification}: Convert all data to Alpaca format, yielding approximately 8.7M samples.
    \item \textbf{Initial filtering}: Remove overly long/short samples and non-English data.
    \item \textbf{Two-stage deduplication}: Apply strict exact Q-matching deduplication, followed by MinHash deduplication (128 hash permutations, threshold 0.7, N-gram 8), producing 570K high-quality instruction samples.
\end{itemize}

The subsequent deduplication further eliminates sample-level overlaps between different leaf nodes. Table~\ref{tab:root_node_list} provides the complete list of the 31 core datasets utilized in our curation process after domain filtering.

\paragraph{Diversity Metric Calculations.}
To rigorously evaluate the semantic span of the curated data and baseline data, we employed two complementary metrics. Let $X = \{x_1, x_2, \dots, x_N\}$ be the set of embeddings for the instructions in the dataset, where $N$ is the sample size, computed using \texttt{Qwen/Qwen3-Embedding-8B}~\cite{qwen3embedding}. We fix the embedding dimensionality to 4096 for all samples.

 \paragraph{(1) Vendi Score (Intrinsic Diversity)} 
    The Vendi Score~\cite{friedman2023vendiscorediversityevaluation} interprets diversity as the effective number of unique semantic clusters. It is calculated based on the eigenvalues of the kernel matrix $K$, where $K_{ij} = k(x_i, x_j)$ represents the similarity between samples (we use the cosine similarity kernel). The Vendi Score is defined as the exponential of the Shannon entropy of the eigenvalues:
    \begin{equation}
        \text{Vendi}(X) = \exp\left( - \sum_{i=1}^{N} \lambda_i \ln \lambda_i \right)
    \end{equation}
    where $\lambda_1, \dots, \lambda_N$ are the normalized eigenvalues of the matrix $K/N$. A higher Vendi Score indicates a dataset with a larger number of effective independent modes. In practice, we compute Vendi using the reference implementation provided by the original authors.\footnote{\url{https://github.com/vertaix/Vendi-Score}}

\paragraph{(2) Centroid Distance (Geometric Dispersion).}
    This metric measures the spatial spread of the data points in the high-dimensional embedding space. We first compute the global centroid $\mu$ of the dataset:
    \begin{equation}
        \mu = \frac{1}{N} \sum_{i=1}^{N} x_i
    \end{equation}
    The Centroid Distance is then defined as the complement of the average cosine similarity between each sample $x_i$ and the centroid $\mu$:
    \begin{equation}
        \text{Dist}_{cent}(X) = 1 - \frac{1}{N} \sum_{i=1}^{N} \frac{x_i \cdot \mu}{\|x_i\| \|\mu\|}
    \end{equation}
    A higher Centroid Distance implies that the samples are widely dispersed around the center, covering a broader semantic region rather than clustering tightly around a single topic.

\onecolumn

\definecolor{SectionColor}{HTML}{D7F6FF} 
\definecolor{HeaderColor}{gray}{0.95}    
\newcommand{\Repo}[1]{{\small\texttt{#1}}} 
\newcommand{\SectionHeader}[1]{%
  \rowcolor{HeaderColor}
  \multicolumn{2}{p{\dimexpr\textwidth-2\tabcolsep}}{%
    \rule{0pt}{2.6ex}\textbf{\textit{Target Benchmark:} #1}\rule[-1.0ex]{0pt}{0pt}%
  } \\%
}
\setlength{\aboverulesep}{0pt}
\setlength{\belowrulesep}{0pt}

\begin{table*}[t]
\centering
\small
\renewcommand{\arraystretch}{1.25} 
\setlength{\tabcolsep}{12pt}      

\caption{Data contamination analysis: leakage ratios of training datasets on various benchmarks.}
\label{tab:data_contamination}

\begin{tabularx}{\textwidth}{
    >{\raggedright\arraybackslash}X       
    >{\raggedleft\arraybackslash}p{4.5cm} 
}
    \toprule
    \textbf{Training Dataset} & \textbf{Contamination Ratio} \\
    \midrule

    \SectionHeader{Omni-MATH}
    \midrule
    \Repo{Skywork/Skywork-OR1-RL-Data} & 96.80\% (4265/4406) \\
    \Repo{agentica-org/DeepScaleR-Preview-Dataset} & 79.48\% (3502/4406) \\
    \Repo{SynthLabsAI/Big-Math-RL-Verified} & 57.97\% (2554/4406) \\
    \Repo{LHL3341/Caco-1.3M} & 37.95\% (1672/4406) \\
    \Repo{a-m-team/AM-Thinking-v1-Distilled} & 28.94\% (1275/4406) \\
    \Repo{PrimeIntellect/SYNTHETIC-2-SFT-verified} & 23.01\% (1014/4406) \\
    \Repo{RabotniKuma/Fast-Math-R1-SFT} & 4.86\% (214/4406) \\
    \Repo{qihoo360/Light-R1-SFTData} & 2.77\% (122/4406) \\
    \midrule

    \SectionHeader{TheoremQA}
    \midrule
    \Repo{garage-bAInd/Open-Platypus} & 70.59\% (564/799) \\
    \Repo{TIGER-Lab/MathInstruct} & 66.96\% (535/799) \\
    \Repo{teknium/OpenHermes-2.5} & 62.33\% (498/799) \\
    \Repo{a-m-team/AM-Thinking-v1-Distilled} & 46.43\% (371/799) \\
    \Repo{open-thoughts/OpenThoughts2-1M} & 8.89\% (71/799) \\
    \Repo{alibaba-pai/OmniThought-0528} & 4.38\% (35/799) \\
    \midrule

    \SectionHeader{LiveCodeBench}
    \midrule
    \Repo{agentica-org/DeepCoder-Preview-Dataset} & 88.12\% (89/101) \\
    \Repo{a-m-team/AM-Thinking-v1-Distilled} & 44.55\% (45/101) \\
    \midrule

    \SectionHeader{TruthfulQA}
    \midrule
    \Repo{openbmb/UltraFeedback} & 99.27\% (811/817) \\
    \midrule

    \SectionHeader{SciBench}
    \midrule
    \Repo{garage-bAInd/Open-Platypus} & 76.23\% (526/690) \\
    \Repo{teknium/OpenHermes-2.5} & 66.09\% (456/690) \\

    \bottomrule
\end{tabularx}
\end{table*}
{\small
\renewcommand{\arraystretch}{1.25} 
\setlength{\tabcolsep}{8pt}     
\newcommand{\Arrow}{\hspace{2pt}\textcolor{gray!80}{$\to$}\hspace{2pt}}
\newcommand{\Num}[1]{\textbf{\textcolor{black!70}{#1.}}}
\definecolor{SectionColor}{HTML}{D7F6FF} 
\definecolor{HeaderColor}{gray}{0.95}   
\setlength{\aboverulesep}{0pt}
\setlength{\belowrulesep}{0pt}

\begin{xltabular}{\textwidth}{
    >{\raggedright\arraybackslash}p{4.0cm}
    >{\raggedright\arraybackslash}X
}
    \caption{Data lineage analysis: tracking the usage of key benchmarks in downstream training datasets.} \label{tab:data_lineage} \\
    \toprule
    \textbf{Training Dataset} & \textbf{Evolutionary Path (Source \Arrow Target)} \\
    \midrule
    \endfirsthead
    \caption[]{Data lineage analysis (Continued).} \\
    \toprule
    \textbf{Training Dataset} & \textbf{Evolutionary Path (Source \Arrow Target)} \\
    \midrule
    \endhead

    \bottomrule
    \multicolumn{2}{r}{\textit{Continued on next page...}} \\
    \endfoot

    \bottomrule
    \endlastfoot


    \SectionHeader{Omni-MATH}
    \midrule
    
    \textbf{\Repo{GAIR/LIMO}} & 
    \Repo{KbsdJames/Omni-MATH} \Arrow 
    \Repo{agentica-org/DeepScaleR-Preview-Dataset} \Arrow 
    \Repo{GAIR/LIMO} \\
    \addlinespace[5pt]

    \textbf{\Repo{LHL3341/Caco-1.3M}} & 
    \Num{1} \Repo{KbsdJames/Omni-MATH} \Arrow \Repo{agentica-org/DeepScaleR-Preview-Dataset} \Arrow \Repo{LHL3341/Caco-1.3M} \newline
    \Num{2} \Repo{KbsdJames/Omni-MATH} \Arrow \Repo{SynthLabsAI/Big-Math-RL-Verified} \Arrow \Repo{LHL3341/Caco-1.3M} \\
    \addlinespace[5pt]

    \textbf{\Repo{PrimeIntellect/\newline SYNTHETIC-2-SFT-verified}} & 
    \Num{1} \Repo{KbsdJames/Omni-MATH} \Arrow \Repo{Skywork/Skywork-OR1-RL-Data} \Arrow \Repo{PrimeIntellect/SYNTHETIC-2-SFT-verified} \newline
    \Num{2} \Repo{KbsdJames/Omni-MATH} \Arrow \Repo{agentica-org/DeepScaleR-Preview-Dataset} \Arrow \Repo{Skywork/Skywork-OR1-RL-Data} \Arrow \Repo{PrimeIntellect/SYNTHETIC-2-SFT-verified} \\
    \addlinespace[5pt]

    \textbf{\Repo{RabotniKuma/\newline Fast-Math-R1-SFT}} & 
    \Num{1} \Repo{KbsdJames/Omni-MATH} \Arrow \Repo{qihoo360/Light-R1-SFTData} \Arrow \Repo{RabotniKuma/Fast-Math-R1-SFT} \newline
    \Num{2} \Repo{KbsdJames/Omni-MATH} \Arrow \Repo{agentica-org/DeepScaleR-Preview-Dataset} \Arrow \Repo{GAIR/LIMO} \Arrow \Repo{qihoo360/Light-R1-SFTData} \Arrow \Repo{RabotniKuma/Fast-Math-R1-SFT} \\
    \addlinespace[5pt]

    \textbf{\Repo{Skywork/\newline Skywork-OR1-RL-Data}} & 
    \Num{1} \Repo{KbsdJames/Omni-MATH} \Arrow \Repo{Skywork/Skywork-OR1-RL-Data} \newline
    \Num{2} \Repo{KbsdJames/Omni-MATH} \Arrow \Repo{agentica-org/DeepScaleR-Preview-Dataset} \Arrow \Repo{Skywork/Skywork-OR1-RL-Data} \\
    \addlinespace[5pt]

    \textbf{\Repo{SynthLabsAI/\newline Big-Math-RL-Verified}} & 
    \Repo{KbsdJames/Omni-MATH} \Arrow \Repo{SynthLabsAI/Big-Math-RL-Verified} \\
    \addlinespace[5pt]

    \textbf{\Repo{a-m-team/\newline AM-Thinking-v1-Distilled}} & 
    \Repo{KbsdJames/Omni-MATH} \Arrow \Repo{SynthLabsAI/Big-Math-RL-Verified} \Arrow \Repo{a-m-team/AM-Thinking-v1-Distilled} \\
    \addlinespace[5pt]

    \textbf{\Repo{agentica-org/\newline DeepScaleR-Preview-Dataset}} & 
    \Repo{KbsdJames/Omni-MATH} \Arrow \Repo{agentica-org/DeepScaleR-Preview-Dataset} \\
    \addlinespace[5pt]

    \textbf{\Repo{qihoo360/\newline Light-R1-SFTData}} & 
    \Num{1} \Repo{KbsdJames/Omni-MATH} \Arrow \Repo{qihoo360/Light-R1-SFTData} \newline
    \Num{2} \Repo{KbsdJames/Omni-MATH} \Arrow \Repo{agentica-org/DeepScaleR-Preview-Dataset} \Arrow \Repo{GAIR/LIMO} \Arrow \Repo{qihoo360/Light-R1-SFTData} \\
    \midrule

    \SectionHeader{TruthfulQA}
    \midrule

    \textbf{\Repo{openbmb/UltraFeedback}} & 
    \Repo{truthfulqa/truthful\_qa} \Arrow \Repo{openbmb/UltraFeedback} \\
    \midrule

    \SectionHeader{LiveCodeBench}
    \midrule

    \textbf{\Repo{a-m-team/\newline AM-Thinking-v1-Distilled}} & 
    \Repo{PrimeIntellect/LiveCodeBench-v5} \Arrow \Repo{agentica-org/DeepCoder-Preview-Dataset} \Arrow \Repo{a-m-team/AM-Thinking-v1-Distilled} \\
    \addlinespace[5pt]

    \textbf{\Repo{agentica-org/\newline DeepCoder-Preview-Dataset}} & 
    \Repo{PrimeIntellect/LiveCodeBench-v5} \Arrow \Repo{agentica-org/DeepCoder-Preview-Dataset} \\
    \midrule

    \SectionHeader{TheoremQA}
    \midrule

    \textbf{\Repo{QizhiPei/ScaleDiff-Math}} & 
    \Repo{TIGER-Lab/TheoremQA} \Arrow \Repo{garage-bAInd/Open-Platypus} \Arrow \Repo{teknium/OpenHermes-2.5} \Arrow \Repo{TIGER-Lab/WebInstructSub} \Arrow \Repo{zwhe99/DeepMath-103K} \Arrow \Repo{QizhiPei/ScaleDiff-Math} \\
    \addlinespace[5pt]

    \textbf{\Repo{TIGER-Lab/MathInstruct}} & 
    \Repo{TIGER-Lab/TheoremQA} \Arrow \Repo{TIGER-Lab/MathInstruct} \\
    \addlinespace[5pt]

    \textbf{\Repo{TIGER-Lab/WebInstruct-CFT}} & 
    \Repo{TIGER-Lab/TheoremQA} \Arrow \Repo{garage-bAInd/Open-Platypus} \Arrow \Repo{teknium/OpenHermes-2.5} \Arrow \Repo{TIGER-Lab/WebInstructSub} \Arrow \Repo{TIGER-Lab/WebInstruct-CFT} \\
    \addlinespace[5pt]

    \textbf{\Repo{TIGER-Lab/WebInstructSub}} & 
    \Repo{TIGER-Lab/TheoremQA} \Arrow \Repo{garage-bAInd/Open-Platypus} \Arrow \Repo{teknium/OpenHermes-2.5} \Arrow \Repo{TIGER-Lab/WebInstructSub} \\
    \addlinespace[5pt]

    \textbf{\Repo{a-m-team/\newline AM-Thinking-v1-Distilled}} & 
    \Repo{TIGER-Lab/TheoremQA} \Arrow \Repo{garage-bAInd/Open-Platypus} \Arrow \Repo{teknium/OpenHermes-2.5} \Arrow \Repo{a-m-team/AM-Thinking-v1-Distilled} \\
    \addlinespace[5pt]

    \textbf{\Repo{alibaba-pai/OmniThought}} & 
    \Num{1} \Repo{TIGER-Lab/TheoremQA} \Arrow \Repo{TIGER-Lab/MathInstruct} \Arrow \Repo{open-thoughts/OpenThoughts2-1M} \Arrow \Repo{alibaba-pai/OmniThought} \newline
    \Num{2} \Repo{TIGER-Lab/TheoremQA} \Arrow \Repo{garage-bAInd/Open-Platypus} \Arrow \Repo{teknium/OpenHermes-2.5} \Arrow \Repo{TIGER-Lab/WebInstructSub} \Arrow \Repo{zwhe99/DeepMath-103K} \Arrow \Repo{alibaba-pai/OmniThought} \\
    \addlinespace[5pt]

    \textbf{\Repo{alibaba-pai/\newline OmniThought-0528}} & 
    \Num{1} \Repo{TIGER-Lab/TheoremQA} \Arrow \Repo{TIGER-Lab/MathInstruct} \Arrow \Repo{open-thoughts/OpenThoughts2-1M} \Arrow \Repo{alibaba-pai/OmniThought} \Arrow \Repo{alibaba-pai/OmniThought-0528} \newline
    \Num{2} \Repo{TIGER-Lab/TheoremQA} \Arrow \Repo{garage-bAInd/Open-Platypus} \Arrow \Repo{teknium/OpenHermes-2.5} \Arrow \Repo{TIGER-Lab/WebInstructSub} \Arrow \Repo{zwhe99/DeepMath-103K} \Arrow \Repo{alibaba-pai/OmniThought} \Arrow \Repo{alibaba-pai/OmniThought-0528} \\
    \addlinespace[5pt]

    \textbf{\Repo{garage-bAInd/Open-Platypus}} & \Repo{TIGER-Lab/TheoremQA} \Arrow \Repo{garage-bAInd/Open-Platypus} \\
    \addlinespace[5pt]

    \textbf{\Repo{open-thoughts/\newline OpenThoughts2-1M}} & \Repo{TIGER-Lab/TheoremQA} \Arrow \Repo{TIGER-Lab/MathInstruct} \Arrow \Repo{open-thoughts/OpenThoughts2-1M} \\
    \addlinespace[5pt]

    \textbf{\Repo{teknium/OpenHermes-2.5}} & \Repo{TIGER-Lab/TheoremQA} \Arrow \Repo{garage-bAInd/Open-Platypus} \Arrow \Repo{teknium/OpenHermes-2.5} \\
    \addlinespace[5pt]

    \textbf{\Repo{zwhe99/DeepMath-103K}} & \Repo{TIGER-Lab/TheoremQA} \Arrow \Repo{garage-bAInd/Open-Platypus} \Arrow \Repo{teknium/OpenHermes-2.5} \Arrow \Repo{TIGER-Lab/WebInstructSub} \Arrow \Repo{zwhe99/DeepMath-103K} \\
    \midrule

    \SectionHeader{SciBench}
    \midrule

    \textbf{\Repo{QizhiPei/ScaleDiff-Math}} & 
    \Repo{xw27/scibench} \Arrow \Repo{garage-bAInd/Open-Platypus} \Arrow \Repo{teknium/OpenHermes-2.5} \Arrow \Repo{TIGER-Lab/WebInstructSub} \Arrow \Repo{zwhe99/DeepMath-103K} \Arrow \Repo{QizhiPei/ScaleDiff-Math} \\
    \addlinespace[5pt]

    \textbf{\Repo{TIGER-Lab/WebInstruct-CFT}} & 
    \Repo{xw27/scibench} \Arrow \Repo{garage-bAInd/Open-Platypus} \Arrow \Repo{teknium/OpenHermes-2.5} \Arrow \Repo{TIGER-Lab/WebInstructSub} \Arrow \Repo{TIGER-Lab/WebInstruct-CFT} \\
    \addlinespace[5pt]

    \textbf{\Repo{TIGER-Lab/WebInstructSub}} & 
    \Repo{xw27/scibench} \Arrow \Repo{garage-bAInd/Open-Platypus} \Arrow \Repo{teknium/OpenHermes-2.5} \Arrow \Repo{TIGER-Lab/WebInstructSub} \\
    \addlinespace[5pt]

    \textbf{\Repo{a-m-team/\newline AM-Thinking-v1-Distilled}} & 
    \Repo{xw27/scibench} \Arrow \Repo{garage-bAInd/Open-Platypus} \Arrow \Repo{teknium/OpenHermes-2.5} \Arrow \Repo{a-m-team/AM-Thinking-v1-Distilled} \\
    \addlinespace[5pt]

    \textbf{\Repo{alibaba-pai/OmniThought}} & 
    \Repo{xw27/scibench} \Arrow \Repo{garage-bAInd/Open-Platypus} \Arrow \Repo{teknium/OpenHermes-2.5} \Arrow \Repo{TIGER-Lab/WebInstructSub} \Arrow \Repo{zwhe99/DeepMath-103K} \Arrow \Repo{alibaba-pai/OmniThought} \\
    \addlinespace[5pt]

    \textbf{\Repo{alibaba-pai/\newline OmniThought-0528}} & 
    \Repo{xw27/scibench} \Arrow \Repo{garage-bAInd/Open-Platypus} \Arrow \Repo{teknium/OpenHermes-2.5} \Arrow \Repo{TIGER-Lab/WebInstructSub} \Arrow \Repo{zwhe99/DeepMath-103K} \Arrow \Repo{alibaba-pai/OmniThought} \Arrow \Repo{alibaba-pai/OmniThought-0528} \\
    \addlinespace[5pt]

    \textbf{\Repo{garage-bAInd/Open-Platypus}} & 
    \Repo{xw27/scibench} \Arrow \Repo{garage-bAInd/Open-Platypus} \\
    \addlinespace[5pt]

    \textbf{\Repo{teknium/OpenHermes-2.5}} & 
    \Repo{xw27/scibench} \Arrow \Repo{garage-bAInd/Open-Platypus} \Arrow \Repo{teknium/OpenHermes-2.5} \\
    \addlinespace[5pt]

    \textbf{\Repo{zwhe99/DeepMath-103K}} & 
    \Repo{xw27/scibench} \Arrow \Repo{garage-bAInd/Open-Platypus} \Arrow \Repo{teknium/OpenHermes-2.5} \Arrow \Repo{TIGER-Lab/WebInstructSub} \Arrow \Repo{zwhe99/DeepMath-103K} \\
\end{xltabular}
}

\begin{table}[t!]
\centering
\small
\renewcommand{\arraystretch}{1.15}
\setlength{\tabcolsep}{4pt}

\caption{List of the 31 Root Nodes ($d_{in}=0$) selected as upstream knowledge anchors. These datasets formed the initial pool for our provenance-based sampling strategy.}
\label{tab:root_node_list}

\begin{tabular}{p{0.48\linewidth} p{0.48\linewidth}}
\toprule
\multicolumn{2}{c}{\textbf{Seed Datasets List}} \\
\midrule
\texttt{sahil2801/CodeAlpaca-20k} & \texttt{deepmind/aqua\_rat} \\
\texttt{glaiveai/glaive-code-assistant-v3} & \texttt{ajibawa-2023/Python-Code-23k-ShareGPT} \\
\texttt{allenai/sciq} & \texttt{mlfoundations-dev/stackexchange\_physics} \\
\texttt{ise-uiuc/Magicoder-OSS-Instruct-75K} & \texttt{allenai/qasc} \\
\texttt{justus27/reasoning-gym-genesys} & \texttt{camel-ai/biology} \\
\texttt{autoprogrammer/nemotron\_science\_lf\_filtered} & \texttt{jondurbin/airoboros-2.1} \\
\texttt{teknium/GPT4-LLM-Cleaned} & \texttt{PrimeIntellect/synthetic-code-understanding} \\
\texttt{mlfoundations-dev/stackexchange\_codegolf} & \texttt{camel-ai/physics} \\
\texttt{PrimeIntellect/real-world-swe-problems} & \texttt{HARP(Human Annotated Reasoning Problems)} \\
\texttt{MatrixStudio/Codeforces-Python-Submission} & \texttt{PrimeIntellect/stackexchange-question-answering} \\
\texttt{hoanganhpham/openr1\_hard} & \texttt{sopen-r1/codeforces} \\
\texttt{avaliev/ChemistryQA} & \texttt{nvidia/OpenScience} \\
\texttt{camel-ai/chemistry} & \texttt{hivaze/LOGIC-701} \\
\texttt{LooksJuicy/ruozhiba} & \texttt{stanfordnlp/web\_questions} \\
\texttt{Multilingual-Multimodal-NLP/McEval-Instruct} & \texttt{di-zhang-fdu/AIME\_1983\_2024} \\
\texttt{Magpie-Align/Magpie-Reasoning-V2-250K-CoT-Llama3} & \\
\bottomrule
\end{tabular}
\end{table}

\end{document}